\documentclass{article} 
\usepackage{iclr2027_conference,times}
\usepackage[letterpaper,left=1.5in,right=1.5in,top=1in,bottom=1in]{geometry}

\usepackage{amsmath,amsfonts,bm}

\def\eqref#1{equation~\ref{#1}}

\def\1{\bm{1}}

\def\vu{{\bm{u}}}
\def\vv{{\bm{v}}}
\def\vw{{\bm{w}}}
\def\vx{{\bm{x}}}
\def\vy{{\bm{y}}}

\def\mW{{\bm{W}}}

\DeclareMathAlphabet{\mathsfit}{\encodingdefault}{\sfdefault}{m}{sl}
\SetMathAlphabet{\mathsfit}{bold}{\encodingdefault}{\sfdefault}{bx}{n}

\usepackage{amsmath}
\usepackage{graphicx}
\usepackage{subcaption}
\newtheorem{theorem}{Theorem}
\usepackage{subcaption}
\usepackage{booktabs}
\usepackage{multirow}
\usepackage{wrapfig}
\usepackage{hyperref}
\hypersetup{hidelinks}
\usepackage{url}
\usepackage{tikz}
\newcommand*\circled[1]{\tikz[baseline=(char.base)]{
            \node[shape=circle,draw,inner sep=0.5pt] (char) {#1};}}

\title{CIPHER-MoE: Balancing Efficiency and Routing Fidelity in Trillion-Scale MoE Training}

\author{\begin{minipage}{\dimexpr\textwidth-2\tabcolsep\relax}
\centering
\normalsize\bfseries
Jing Li\textsuperscript{1}\thanks{Equal contribution.}, Jian Meng\textsuperscript{2,*}\thanks{Corresponding author: \texttt{jm2787@cornell.edu}}, Yingmeng Gao, Suming Qiu\\[3pt]
Linyuan Qiu, Dongfang Li\textsuperscript{3,4}, Baotian Hu\textsuperscript{3,4}, Binfan Zheng\\[3pt]
Rongqian Zhao, Weijian Sun, Xin Chen\\[7pt]
\normalsize\normalfont
\textsuperscript{1}Tongji University\quad \textsuperscript{2}Cornell University\\
\textsuperscript{3}Harbin Institute of Technology, Shenzhen\\
\textsuperscript{4}AI Training Platform Team, Shenzhen Loop Area Institute
\end{minipage}}

\iclrfinalcopy 
\begin{document}

\maketitle

\begin{abstract}
Mixture-of-Experts~(MoE) has been widely adopted in recent large language model~(LLM) architectures. However, scaling up MoE in LLM training introduces system-level challenges on training, where non-uniform token routing can lead to highly imbalanced workloads across experts and devices, further destabilizing the training process. With trillion-scale LLMs, imbalanced expert workloads further amplify the resource cost of MoE training, resulting in degraded training efficiency and hardware utilization for underloaded experts, while hot experts require additional resources to accommodate excessive workloads. Recent studies address imbalanced MoE training through intricate parallelism strategies or resource reallocation. However, these system-level approaches often introduce additional resource requirements and considerable orchestration complexity, which become increasingly difficult to afford when training trillion-parameter LLMs under constrained computational resources. 
This work introduces CIPHER-MoE, which mitigates MoE workload imbalance while keeping the router's token-side Top-K selection unchanged. CIPHER-MoE applies affinity-aware Expert-to-Token filtering with explicit capacity control to reduce hotspot expert workloads without additional hardware resources or complex runtime design. The proposed method has been evaluated on large-scale MoE models, including DeepSeek-V4-Pro, showing up to 64.9 percentage points Top-1 expert workload reduction and 1.10$\times$--1.94$\times$ training acceleration, while preserving the training quality. The source code will be released soon.
\end{abstract}

\section{Introduction}
Large language models~(LLMs) have become foundational components of modern AI systems~\citep{bommasani2021opportunities}. The continued effectiveness of scaling laws has demonstrated that increasing model capacity can substantially improve performance~\citep{kaplan2020scaling}. However, naively scaling dense Transformers toward the trillion-parameter regime incurs prohibitively high computational and memory costs, motivating the adoption of Mixture-of-Experts~(MoE) architectures~\citep{fedus2022switch}. In particular, MoE decouples model capacity from per-token computation by replacing the dense feed-forward network~(FFN) with multiple sparsely activated experts and dynamically routing each token to only a small subset of them, leading to a balanced tradeoff between scalability and computational efficiency~\citep{shazeer2017outrageously}.

Recent studies have demonstrated the strong potential of domain-specific supervised fine-tuning~(SFT) and reinforcement learning~(RL) for advancing LLM post-training~\citep{guo2025deepseek}. Training MoE models introduces substantial system-level challenges due to workload imbalance, which degrades communication efficiency, hardware utilization, and training stability~\citep{jin2026relibra}. 
The inherently non-uniform data distribution across domains and tasks can induce highly skewed token routing, resulting in uneven workloads across different experts. 
For the expensive trillion-parameter LLMs, cross-rank expert parallelism is almost mandatory for efficient training. However, 
the lopsided computation impedes efficient distributed training with unbalanced communication and computation~\citep{rajbhandari2022deepspeed}. Moreover, the overloaded experts may incur excessive memory consumption and instantly trigger out-of-memory failures on large-scale computing clusters, ultimately undermining training efficiency and stability~\citep{nguyen2026least}. 


\begin{figure*}
    \centering
    \includegraphics[width=0.7\linewidth]{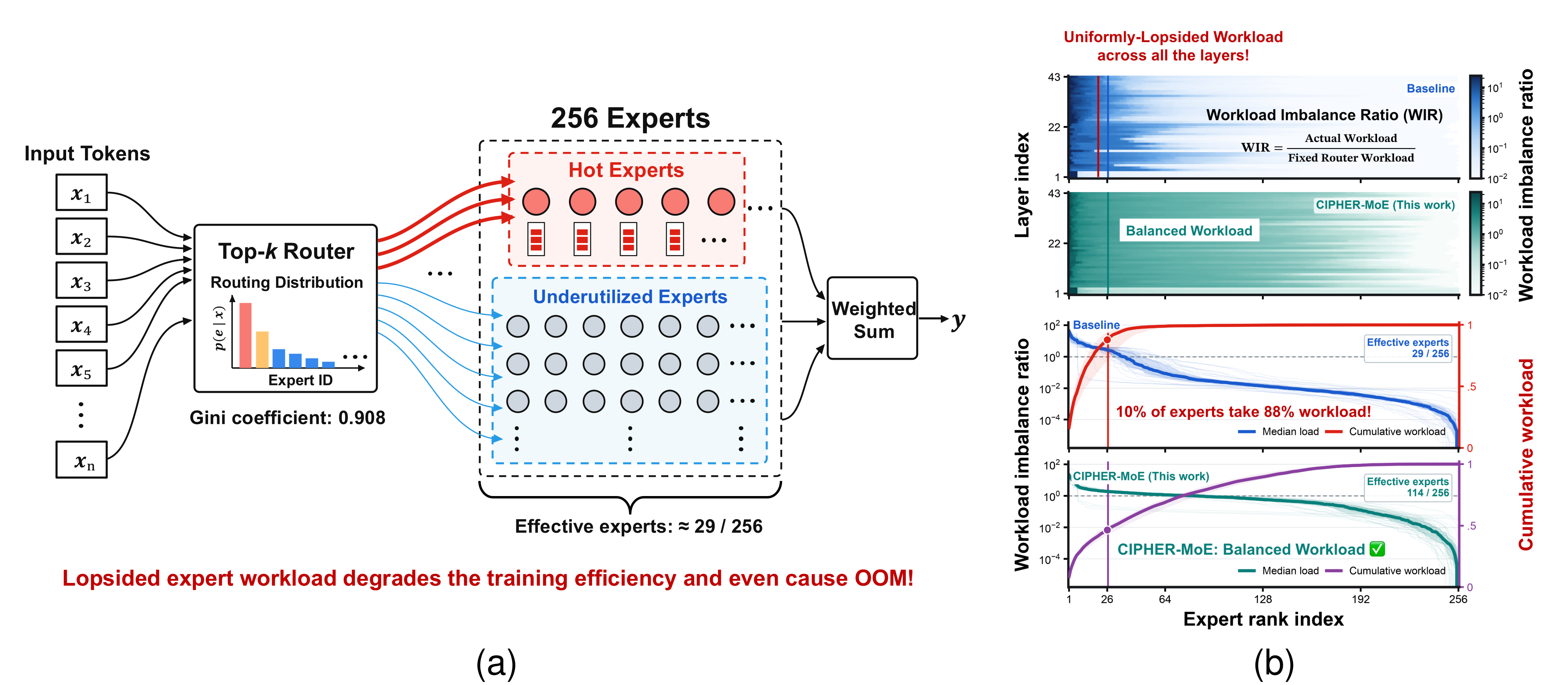}
    \vspace{-1pt}
    \caption{Expert workload imbalance in DeepSeek-V4-Flash. (a) Illustration of the MoE architecture with data-dependent routing, which leads to the imbalanced workload. (b) Lopsided workload statistics across all layers of the DeepSeek-V4-Flash model. With the domain-specific dataset, 10\% of the experts consume over 88\% routing workload, leading to degraded training efficiency.}
    \label{fig:placeholder}
\end{figure*}

Recent studies on efficient MoE training are two-fold; system-level approaches redistribute workloads through dynamic expert migration or replication~\citep{nie2023flexmoe}. Meanwhile, MoE-specific kernels optimize expert-parallel execution through aggressive communication-computation pipeline design~\citep{zhang2025comet}. 
At the algorithm level, recent studies investigate global-batch balancing, routing regularization, or transfer the pre-trained routing strategy to fresh training~\citep{lewis2021base}. 
However, system-level approaches largely accommodate routing imbalance after it arises, requiring additional expert replicas, collective communication, or specialized runtime scheduling, whose overhead and implementation complexity become increasingly pronounced with trillion-parameter LLMs training. 
On the other hand, 
regularization-based routing penalty failed to handle the instantaneous memory peak~\citep{fedus2022switch}, while the heuristic batch balancing or experience-based routing distillation exhibits poor scalability to the trillion-parameter MoE models. 
Given the research gaps in existing algorithm and system-level approaches for MoE LLM training, this work aims to investigate the following problem:

\begin{center}
    \textit{How to achieve efficient and stable training for trillion-parameter MoE LLMs without intricate system design or heuristic-based routing?}
\end{center}

In this work, we propose \texttt{CIPHER-MoE}, a \textit{plug-and-play} algorithm for stable-and-efficient trillion-parameter MoE model training. 
Although the fixed routing strategy ensures the perfect load balance with stabilized training process, the reduced expert specialization leads to the suboptimal intelligence in the resulting trained model~\citep{dikkala2023benefits}. 
On the other hand, freely-exploited expert routing amplifies the workload imbalance~\citep{guo2026advancing}, especially with domain-specific post-training phase. In contrast, the proposed \texttt{CIPHER-MoE} tackles the problem by investigating a bounded learning capacity of different experts and proves the feasibility of dropping tokens from expert training. 
Different from heuristic token dropping, this work first reveals the relationship between optimal token budget and training efficiency for trillion-parameter MoE training. We validate the proposed method via both theoretical analysis and empirical experiments across multiple large-scale MoE models~(from 284B to 1.6T). The major contributions of this paper are:
\begin{itemize}
    \item \textbf{Simplicity}. CIPHER-MoE directly exploits the bounded token budget for MoE training, achieving high efficiency while preserving the LLM training quality.
    \item \textbf{Theoretical Foundation}. CIPHER-MoE derives a sufficient expert capacity for similarity-based admission and a necessary capacity for natural-order admission. 
    \item \textbf{Scalability}. CIPHER-MoE scales effectively across different model sizes, architectures, and datasets. Across DeepSeek-V3, DeepSeek-V4-Flash, GLM-5, and DeepSeek-V4-Pro, achieving  $\mathbf{1.10\times}$--$\mathbf{1.94\times}$ measured training-time speedups.
\end{itemize}

\section{Related Work}
\label{related_work}
\paragraph{System-level workload balancing and efficient MoE execution.} Workload imbalance is a major bottleneck for efficient distributed MoE training~\citep{hwang2023tutel}, where skewed token routing creates expert stragglers, uneven communication, and poor accelerator utilization. System-level work mitigates this imbalance by adapting the mapping between experts and hardware. FasterMoE and FlexMoE replicate, resize, or migrate experts according to runtime load~\citep{he2022fastermoe,nie2023flexmoe}, while some works combine balancing with parameter offloading, decoupled parameter/optimizer-state placement, or sharded expert re-layout~\citep{3692070.3693045,skiadopoulos2025accelerating,liu2026laer}. In parallel, another line of work reduces the overhead of sparse and irregular MoE computation. Tutel and PipeMoE~\citep{hwang2023tutel,shi2023pipemoe} optimize expert-parallel execution through adaptive parallelism and communication--computation pipelining; MegaBlocks~\citep{gale2023megablocks} uses block-sparse kernels to process variable expert workloads without padding or token dropping; MegaScale-MoE~\citep{jin2026megascale} jointly optimizes parallelism, communication--computation overlap, and communication compression. While these techniques improve MoE execution efficiency, they operate after routing and rely on expert replication, parameter migration or re-layout, runtime profiling, and specialized communication scheduling, increasing system complexity at scale.

\paragraph{Algorithmic load balancing and expert specialization.}
Routing offers an alternative approach to load balancing. GShard~\citep{lepikhin2020gshard} and Switch Transformer~\citep{fedus2022switch} use auxiliary load-balancing objectives to encourage uniform expert utilization. However, overly strong local balancing can interfere with the language-modeling objective and suppress expert specialization. Loss-Free Balancing~\citep{wang2024auxiliary}, adopted by DeepSeek-V3~\citep{liu2024deepseek}, adjusts per-expert routing biases based on historical loads without introducing auxiliary-loss gradients. Global-Batch Load Balancing~\citep{qiu2025demons} aggregates global statistics to preserve domain specialization. Related work further highlights the tension between strict balancing and expert differentiation~\citep{dai2024deepseekmoe}, while $\phi$-balancing~\citep{chen2026phi} formulates expert utilization as a population-level objective rather than relying on noisy mini-batch statistics. These methods improve average utilization through auxiliary objectives, historical statistics, or batch-level assignments rather than bounding instantaneous loads, although peak memory and latency depend on the most heavily loaded expert.

\paragraph{Capacity-aware token--expert assignment.}
Several works are closely related to our capacity-aware perspective. LocMoE~\citep{10.24963/ijcai.2024/705} jointly considers workload balance and communication locality, deriving a minimum expert capacity from the angular relationship between token representations and gating weights. This result motivates capacity-aware routing by showing that expert capacity need not be overprovisioned to preserve model quality. Expert Choice Routing~\citep{zhou2022mixture} instead fixes per-expert capacity by reversing token-to-expert assignment. These approaches show that expert-side token selection can control workloads, but either impose fixed expert-centric assignments or couple capacity optimization with specific locality and routing assumptions. In contrast, \texttt{CIPHER-MoE} analyzes expert capacity for large-scale training and establishes a sufficient capacity bound that enables workload-aware \emph{bidirectional} token--expert assignment. Tokens first identify preferred experts, which then select the highest-affinity tokens within the derived budget, while excess assignments are rerouted. This design preserves token--expert affinity and controls hotspot workloads without expert replication, parameter migration, or complex runtime orchestration. We evaluate CIPHER-MoE on MoE models ranging from 284B to 1.6T parameters.

\section{Proposed Method: CIPHER-MoE}
\label{method}
\subsection{Preliminaries of Mixture-of-Expert training}
\label{sec:moe_preliminary}

\paragraph{Sparse expert computation and training.}
An MoE layer replaces the dense FFN with $E$ expert networks $\{f_e(\cdot;\theta_e)\}_{e=1}^{E}$ and a router that activates only $K\ll E$ experts for each token~\citep{fedus2022switch}. Consider $N$ token representations $\{\vx_i\}_{i=1}^{N}$, with $\vx_i\in\mathbb{R}^{d}$, processed by one expert-parallel group in a training micro-batch~(the layer index is omitted for clarity). 
Let $\mW_r\in\mathbb{R}^{E\times d}$ denote the router parameters and $\vw_e^\top$ its $e$-th row. A conventional softmax router computes the token--expert affinity $p_{i,e}$ and selects the corresponding Top-$K$ expert set $\mathcal{S}_i$:
\begin{equation}
    s_{i,e}=\vw_e^\top\vx_i,\qquad
    p_{i,e}=\frac{\exp(s_{i,e})}{\sum_{j=1}^{E}\exp(s_{i,j})},\qquad
    \mathcal{S}_i=\operatorname{TopK}_{e\in\{1,\ldots,E\}}(p_{i,e}).
    \label{eq:moe_routing}
\end{equation}
Here, $s_{i,e}$ is the routing logit, and $\mathcal{S}_i$ contains the indices of the $K$ selected experts. Define the assignment indicator $a_{i,e}=\mathbf{1}[e\in\mathcal{S}_i]$ and the sparse gate $g_{i,e}=a_{i,e}p_{i,e}$. The routed MoE output is
\begin{equation}
    \vy_i=\sum_{e=1}^{E}g_{i,e}f_e(\vx_i;\theta_e),
    \qquad \sum_{e=1}^{E}a_{i,e}=K.
    \label{eq:moe_output}
\end{equation}
Training optimizes the task loss $\mathcal{L}_{\mathrm{task}}$ over the expert parameters and router parameters. Each token contributes task gradients through its selected expert paths. Consequently, increasing $E$ expands the available model capacity while the expert computation per token remains governed by $K$.

\begin{table}[h!]

\resizebox{\linewidth}{!}{\begin{tabular}{ccccccc}
\toprule
\textbf{Models}            & \textbf{Method} & \textbf{Latency / Iteration~(sec)}               & \textbf{NL4Opt} & \textbf{OptiBench} & \textbf{Bench4Opt Feasi.} & \textbf{Bench4Opt OR} \\ \midrule
DeepSeek-V4-Flash & Fixed Router &   \textbf{28.30}~(\textbf{1.00}$\times$)      & 88.93  & 66.17     & 70.93            & 48.73        \\ \midrule
DeepSeek-V4-Flash & Vanilla MoE Training & 38.69~(1.37$\times$) & \textbf{93.08}  & \textbf{68.00}     & \textbf{71.51}            & \textbf{51.02}        \\ \midrule \midrule
DeepSeek-V4-Pro & Fixed Router &   \textbf{24.68}~(\textbf{1.00}$\times$)      & 89.27  & 64.67     & 74.42            & 54.31        \\ \midrule
DeepSeek-V4-Pro & Vanilla MoE Training & OOM & -  & -     & -            & -        \\ \bottomrule
\end{tabular}}
\caption{Post-training SFT of the DeepSeek-V4-Flash and DeepSeek-V4-Pro model on linear programming and operations research~(OR) dataset. OR capabilities evaluated on NL4Opt~\citep{ramamonjison2023nl4opt}, OptiBench~\citep{yang2025optibench}, Bench4Opt-Feasi.~\citep{wang2025orgeval}, and Bench4Opt-OR~\citep{wang2025orgeval}. Fixed router exhibits suboptimal performance compared to the vanilla MoE training due to the homogenized expert training. Detailed training setup and resource configurations are summarized in the Appendix~\ref{app:configuration}.}
\label{tbl:fix-router}
\end{table}


\paragraph{Fixed Router and homogenized experts.}
With expert parallelism~(EP), experts are distributed across $R$ devices, with $\mathcal{E}_r$ denoting the set hosted on device $r$. 
Given the unpredictable routing for each incoming token, \emph{fixed routing} is introduced to enforce the deterministic workload balancing between different experts. 
For a local routing tensor with $M$ tokens, the $j$-th expert assigned to token position $t$ is
\begin{equation}
    e^{\mathrm{fix}}_{t,j}=1+\bigl((tK+j)\bmod E\bigr),
    \qquad t=0,\ldots,M-1,\quad j=0,\ldots,K-1,
    \label{eq:fixed_router}
\end{equation}
For $K\leq E$, each token is dispatched to $K$ \textbf{distinct} experts, and each expert receives either $\lfloor MK/E\rfloor$ or $\lceil MK/E\rceil$ assignments from this local tensor. 

The predictable routing strategy ensures the stable and balanced training workload for MoE model at the cost of removing affinity-based expert selection. 
Semantically unrelated tokens occupying the same positions across micro-batches are repeatedly dispatched to the same expert combinations, potentially weakening expert differentiation, leading to expert homogenization.
As a result, the trained model exhibits suboptimal domain specialization with homogenized experts~\citep{qiu2025demons}. 
We compare the routing strategy with large-scale MoE model training with the results in Table~\ref{tbl:fix-router}.
The standard training paradigm with freely-explored expert workload reaches the ideal performance, with the cost of the degraded training efficiency. 
With the scaled computing resources, DeepSeek-V4-Pro~(1.6T parameters) even failed before the very first iteration of training with the error of out-of-memory~(OOM), where the freely-explored expert routing explodes the imbalanced workload, resulting in the instant OOM error.

\subsection{CIPHER-MoE: Bidirectional Expert Workload Balancing}
\label{sec:cipher_moe}

\paragraph{Expert-token similarity.}
Given the token hidden state $\vx_i$ and expert gating vector $\vw_e$, their cosine similarity is defined as
\begin{equation}
    c_{i,e}=\frac{\vw_e^\top\vx_i}{\|\vw_e\|_2\|\vx_i\|_2}.
    \label{eq:cipher_cosine}
\end{equation}

For each expert $e$, CIPHER-MoE uses $c_{i,e}$ to rank the candidate tokens in $\mathcal P_e$. The similarity threshold $\tau\in[-1,1]$ excludes tokens with low similarity, while the expert capacity $C_e$ limits the number of tokens processed in each micro-batch.

Specifically, let $\mathcal P_e^\tau=\{i\in\mathcal P_e:c_{i,e}\geq\tau\}$ contain the tokens that satisfy the similarity threshold. If more than $C_e$ tokens satisfy the threshold, the expert retains the $C_e$ tokens with the highest cosine similarity; otherwise, it retains all tokens in $\mathcal P_e^\tau$:
\begin{equation}
    \mathcal T_e^{\mathrm S}
    =\operatorname{Top}_{\min\{C_e,|\mathcal P_e^\tau|\}}
    \bigl(\{c_{i,e}:i\in\mathcal P_e^\tau\}\bigr),
    \qquad b_{i,e}^{\mathrm S}=\mathbf1[i\in\mathcal T_e^{\mathrm S}],
    \label{eq:cipher_strict_selection}
\end{equation}
where $\operatorname{Top}_q$ returns the indices of the $q$ tokens with the highest similarity. The budget $C_e$ can vary with the layer, expert, and training stage, while remaining fixed for each micro-batch. 
Furthermore, the rejection by one expert does not remove a token from its other selected experts, which will be discussed in Section~\ref{sec:strict_vs_reroute}.

Although similarity-based token selection directly controls expert workload, two critical questions require further investigation:
\circled{1} Under what conditions can similarity-based selection reduce the required expert capacity while retaining informative tokens?
\circled{2} How should the routing weights of rejected assignments be handled?

\subsubsection{Theoretical analysis of similarity-based token selection}
\label{sec:cipher_capacity}

To address the first question, we examine why selecting tokens by similarity can require less expert capacity than processing them in their original order. Fix an expert $e$ and its $N_e$ candidate tokens. A token is called \emph{discriminative} if it carries the expert-specific signal described by our assumptions. Let $A_e$ denote the event that the candidate pool contains at least one such token. We compare the two selection policies on the same pool, with the same target:

\begin{equation}
    \Pr(\text{at least one discriminative token is retained}\mid A_e)
    \geq1-\varepsilon,
    \qquad 0<\varepsilon<1.
    \label{eq:cipher_retention_target}
\end{equation}

Natural-order selection processes the first $C$ candidates, whereas CIPHER-MoE ranks the candidates by similarity before allocating its $C$ expert-execution slots.

\paragraph{Assumptions.}
Our analysis uses the assumptions of A1--A3 in Appendix~\ref{sec:cipher_assumptions}. First, candidates are independent draws from an expert's exposure distribution, with signal availability $\Pr(A_e)\geq a_0>0$. Under the spherical reference model, discriminative tokens have normalized affinity at least $\delta\in(0,1)$, and expert exposure has probability at least $1/(\beta E)$, where $\beta\geq1$. Second, conditional on $A_e$, a chosen discriminative token survives candidate filtering and thresholding with probability at least $1-\eta$, where $0\leq\eta<\varepsilon$. Third, after conditioning on its survival and previously examined candidates, the probabilities that competing tokens rank ahead of this token have sum at most $\mu$. Thus, $\eta$ measures the risk of excluding the signal before ranking, while $\mu$ controls the capacity needed to retain it after ranking. The signal margin $\delta$ is distinct from the threshold $\tau$.

\begin{theorem}[Similarity-based Capacity Reduction]
\label{thm:cipher_capacity}
Under A1--A3, with $d\geq2$, define
\begin{equation}
\begin{aligned}
    \ell&=\log\frac{1}{\varepsilon-\eta},
    &G(d,\delta)&=-\frac{d-1}{2}\log(1-\delta^2),\\
    B_e&=\min\left\{N_e,
    \left\lceil1+\mu+\frac{\ell}{3}
    +\sqrt{2\mu\ell+\frac{\ell^2}{9}}\right\rceil\right\}.
\end{aligned}
\label{eq:cipher_sufficient_budget}
\end{equation}
Let $C^*_{\mathrm{natural}}$ and $C^*_{\mathrm{CIPHER}}$ be the minimum capacities required by the respective policies to satisfy Eq.~\ref{eq:cipher_retention_target}. Then
\begin{equation}
    C^*_{\mathrm{CIPHER}}\leq B_e,
    \qquad
    C^*_{\mathrm{natural}}
    \geq\frac{(1-\varepsilon)a_0}{\beta E}e^{G(d,\delta)}.
    \label{eq:cipher_capacity_separation}
\end{equation}
Consequently, a factor-$\kappa$ reduction, $C^*_{\mathrm{CIPHER}}\leq C^*_{\mathrm{natural}}/\kappa$ for $\kappa>1$, is guaranteed whenever the natural-order lower bound is at least $\kappa B_e$.
\end{theorem}

\paragraph{Key proof steps.}
\circled{1}~\textbf{Signal scarcity.} Spherical concentration bounds the discriminative-token probability by $\pi_e\leq\beta E e^{-G(d,\delta)}$. Processing the first $C$ tokens retains a discriminative token with conditional probability at most $C\pi_e/a_0$. Reaching the target therefore requires $C\geq(1-\varepsilon)a_0/\pi_e$, giving the natural-order lower bound.
\circled{2}~\textbf{Similarity ranking.} Let $S$ count candidates ranked ahead of the surviving discriminative token. Its rank is $S+1$. A conditional concentration bound on $S$, using A3, gives the sufficient capacity $B_e$ with ranking-failure probability at most $\varepsilon-\eta$. Including filtering failures keeps the total failure probability at most $\varepsilon$.
\circled{3}~\textbf{Capacity comparison.} Comparing CIPHER-MoE's sufficient capacity with natural-order selection's necessary capacity establishes the reduction. The full proof and explicit dimension condition are given in Appendix~\ref{sec:cipher_capacity_theorem}.

\begin{figure}[t!]
    \centering
    \includegraphics[width=\linewidth]{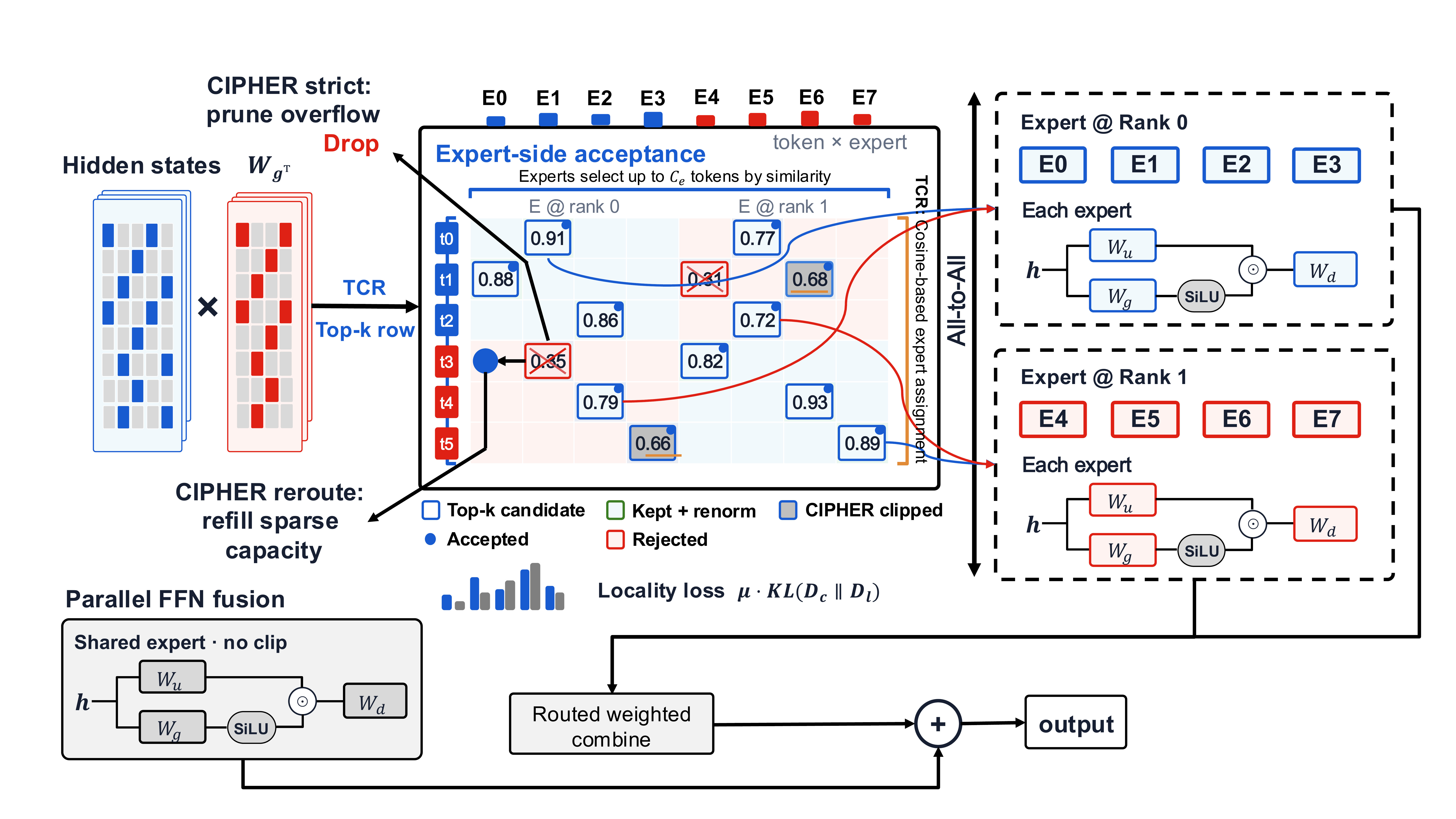}
    \caption{The overall design of the proposed CIPHER-MoE. The incoming tokens start with the normal Top-K expert selection, followed by the proposed expert-side token drop and reroute.}
    \label{fig:cipher_moe_example}
\end{figure}

The result explains the benefit of similarity-based selection: \textbf{An expert can retain an informative token without processing all candidates that precede it in the original order.} Under the stated assumptions, $C_e\geq B_e$ is sufficient for signal retention. This is a requirement on the available capacity, not a minimum number of tokens that must actually be processed; equation~\ref{eq:cipher_strict_selection} still caps the workload at $C_e$. The theorem does not guarantee unchanged training loss or final model quality, and it does not recover signals excluded by an overly restrictive $\tau$. Stability conditions and output-fidelity bounds are provided in Appendices~\ref{sec:cipher_stability} and~\ref{sec:cipher_fidelity}. 

\subsubsection{Strict Token Drop vs. Token--Expert Reroute}
\label{sec:strict_vs_reroute}
Given the theoretical analysis of bounded expert capacity, handling rejected token--expert assignments becomes critical for both the efficiency and quality of LLM training.
\texttt{Strict Token Drop} removes each rejected assignment and sets its routing weight to zero, while retaining the token's other accepted expert paths. \texttt{Token--Expert Rerout}e instead redirects the rejected assignment to another candidate expert that satisfies the similarity threshold and has available capacity.
Both strategies apply \textbf{only} to routed experts. Shared experts continue to process all incoming tokens, and rejecting an expert assignment does not remove the token from the training sequence or the task loss.
Figure~\ref{fig:cipher_moe_example} illustrates these two mechanisms within CIPHER-MoE. 

We evaluate the impact of \textbf{both} strategies on large-sized~(MoE models $>$250B) with domain-specific training tasks. Surprisingly, rerouting tokens to other experts leads to sub-optimal performance compared to the strict token dropping, as shown in Table~\ref{tab:eval-domain}.
Although rerouting preserves more expert computation and distributes overflow to experts with spare capacity and sufficient affinity, the additional computation does not necessarily improve training quality.

For domain-specific training, the token--expert rerouting shifts the token distributions on which those experts specialize. 
Even threshold-qualified tokens can induce updates that interfere with an expert's existing specialization, since cosine affinity alone does not constrain the direction of the task gradient.
Strict token dropping avoids the additional assignments and gradient interference, which allows each expert to train on the highest-affinity candidates retained within its budget. 

\subsubsection{Training Objective of CIPHER-MoE}

On top of the proposed token selection during MoE training, we adopt the Locality loss proposed by LocMoE~\citep{10.24963/ijcai.2024/705}, which minimizes the KL divergence between the current expert-assignment distribution and the ideal distribution target corresponding to all the locally-deployed experts~(e.g., within the same EP rank), softly encouraging tokens to be routed to local experts while preserving load balance. Overall, the training objective of CIPHER-MoE is formulated as:

\begin{align}
\mathcal{L}_{\mathrm{CIPHER}}^{\rho}(\Theta)
&=
-\frac{1}{|\mathcal{U}|}\sum_{t\in\mathcal{U}}\log p_{\Theta}^{\rho}\left(z_t \mid z_{<t}\right)+\lambda_{\mathrm{loc}}\sum_{\ell\in\mathcal{M}}\mathcal{L}_{\mathrm{loc}}^{\rho,\ell}(\Theta),
\\
\text{where}\qquad\mathcal{L}_{\mathrm{loc}}^{\rho,\ell}
&=-\int
D_{c}^{\rho,\ell}(x)\log\left(\frac{D_{l}^{\rho,\ell}(x)}{D_{c}^{\rho,\ell}(x)}\right)\,\mathrm{d}x
=
\operatorname{KL}\left(D_{c}^{\rho,\ell}\,\middle\|\,D_{l}^{\rho,\ell}\right).
\end{align}

where \(D_c\) is the current distribution of expert assignments for tokens in the batch, and \(D_l\) is the hypothetical fully localized assignment distribution. The loss pushes the actual routing distribution toward one where tokens are preferentially processed by experts that are local to their current communication group of expert parallelism~(EP). $M^{\rho,\ell}$ can be determined by the aforementioned strict or reroute strategies, where $\rho \in \{\mathrm{strict}, \mathrm{reroute}\}$.
Other auxiliary loss terms~(e.g., for MTP head and sparse attention indexer) are  omitted for better clarity.

\section{Experimental Results}
\label{sec:exp}
We validate the proposed CIPHER-MoE algorithm on MoE models ranging from 284B to 1.6T parameters, including DeepSeek-V4-Pro (1.6T)~\citep{deepseekai2026deepseekv4}, DeepSeek-V4-Flash (284B)~\citep{deepseekai2026deepseekv4}, GLM-5 (744B)~\citep{glm5team2026glm5vibecodingagentic}, and DeepSeek-V3 (671B)~\citep{liu2024deepseek}.
The selected models are trained on domain-specific datasets including OR-based mathematical reasoning, Medical knowledge MedMCQA~\citep{pmlr-v174-pal22a}, and Social interactions SIQA~\citep{sap2019social}. 
After training the model on domain-specific datasets, we validate the robustness of the trained model on the general-knowledge datasets, as presented in Appendix~\ref{sec:general_benchmark}. 
We would like to highlight the fact that this work is not deliberately picking the skewed data distributions to enforce the lopsided expert workloads. In fact, the domain-specific fine-tuning on powerful foundation LLMs is widely employed in practical scenarios. 
The detailed training configurations are summarized in Appendix~\ref{app:configuration}. 

While most existing works are validated on small-sized MoE models rather than the mega-sized model, this work uses the vanilla MoE training~(provided by the official recipes from Megatron), Fixed-routing enabled MoE training, and  LocMoE~\citep{10.24963/ijcai.2024/705} as the comparison counterpart.

\begin{table*}[h!]
\centering
\caption{Evaluation results and training speedup of domain-specific SFT. The proposed CIPHER-MoE achieves the optimal speedup and quality.}
\label{tab:eval-domain}
\scriptsize
\setlength{\tabcolsep}{3.5pt}
\renewcommand{\arraystretch}{1.08}
\resizebox{\textwidth}{!}{%
\begin{tabular}{llrrrrrr|rr|rr}
\toprule
\textbf{Model} & \textbf{Method}
& \multicolumn{6}{c|}{\textbf{Operations Research}}
& \multicolumn{2}{c|}{\textbf{MedMCQA}}
& \multicolumn{2}{c}{\textbf{SIQA}} \\
\cmidrule(lr){3-8}\cmidrule(lr){9-10}\cmidrule(lr){11-12}
& & \textbf{NL4Opt} & \textbf{OptiBench} & \textbf{B4O-Feas.} & \textbf{B4O-OR} & \textbf{W. Avg.}
& $\mathbf{S}_{\mathrm{train}}$
& \textbf{Acc.} & $\mathbf{S}_{\mathrm{train}}$
& \textbf{Acc.} & $\mathbf{S}_{\mathrm{train}}$ \\
\midrule
\textbf{DeepSeek-V4-Flash}
& Vanilla MoE Training~(Baseline)
& 93.08 & \underline{68.00} & \underline{71.51} & 51.02 & 69.09
& -
& 76.60 & -
& 51.84 & - \\

& CIPHER-strict
& 93.77 & 66.67 & 70.64 & 51.27 & 68.59
& \textbf{1.42$\times$}
& \textbf{77.36} & \textbf{1.50$\times$}
& 54.81 & \textbf{1.58$\times$} \\

& CIPHER-reroute
& 90.66 & 66.33 & 69.77 & 51.27 & 67.73
& \underline{1.38$\times$}
& 76.79 & \underline{1.46$\times$}
& \textbf{57.83} & \underline{1.52$\times$} \\

& LocMoE ($\tau=0.005$)
& \underline{94.12} & \textbf{68.17} & 66.28 & 50.76 & 68.16
& 1.17$\times$
& -- & --
& -- & -- \\

& LocMoE ($\tau=0.01$)
& \textbf{94.81} & 67.67 & 70.06 & \textbf{53.05} & \textbf{69.46}
& 1.12$\times$
& \underline{77.22} & 1.12$\times$
& \underline{55.94} & 1.19$\times$ \\

& LocMoE ($\tau=0.02$)
& 93.08 & 67.50 & \textbf{72.67} & 51.52 & \underline{69.27}
& 1.12$\times$
& -- & --
& -- & -- \\

& LocMoE ($\tau=0.03$)
& 93.77 & 65.67 & 70.93 & \underline{52.79} & 68.65
& 1.10$\times$
& -- & --
& -- & -- \\
\midrule
\textbf{DeepSeek-V3}
& Vanilla MoE Training~(Baseline)
& 83.39 & 60.17 & 58.43 & 46.45 & 60.60
& -
& 72.32 & -
& \textbf{53.63} & - \\

& CIPHER-strict
& \underline{83.74} & 59.83 & \textbf{61.34} & \textbf{47.46} & 61.40
& \textbf{1.94$\times$}
& \underline{72.84} & \textbf{1.84$\times$}
& 52.00 & \textbf{1.78$\times$} \\

& CIPHER-reroute
& 82.70 & \underline{61.50} & \underline{61.05} & \underline{47.20}
& \textbf{61.71}
& 1.68$\times$
& \textbf{73.44} & \underline{1.62$\times$}
& 52.25 & \underline{1.56$\times$} \\

& LocMoE ($\tau=0.01$)
& \textbf{84.08} & \textbf{62.00} & 59.30 & 45.94 & \underline{61.46}
& \underline{1.75$\times$}
& 72.70 & 1.62$\times$
& \underline{53.07} & 1.50$\times$ \\
\midrule
\textbf{GLM-5}
& Vanilla MoE Training~(Baseline)
& 85.47 & \underline{64.67} & \textbf{61.05} & 45.94 & \underline{63.06}
& -
& \underline{80.66} & -
& \textbf{55.83} & - \\

& CIPHER-strict
& \textbf{90.66} & \textbf{65.67} & 57.56 & \underline{46.95} & \textbf{63.86}
& \textbf{1.14$\times$}
& \textbf{81.31} & \textbf{1.14$\times$}
& \underline{55.02} & \textbf{1.13$\times$} \\

& CIPHER-reroute
& 88.58 & 63.33 & \underline{58.14} & \underline{46.95} & 62.75
& 1.10$\times$
& 80.56 & \underline{1.11$\times$}
& 53.63 & 1.10$\times$ \\

& LocMoE ($\tau=0.01$)
& \underline{88.93} & 62.33 & 57.56 & \textbf{47.72} & 62.51
& \underline{1.12$\times$}
& 77.77 & 1.11$\times$
& 50.15 & \underline{1.11$\times$} \\
\bottomrule
\end{tabular}%
}
\end{table*}


\subsection{Training Efficiency and Memory Usage}

\begin{figure*}[!t]
    \centering
    \vspace{-10pt}
    \includegraphics[width=\linewidth]{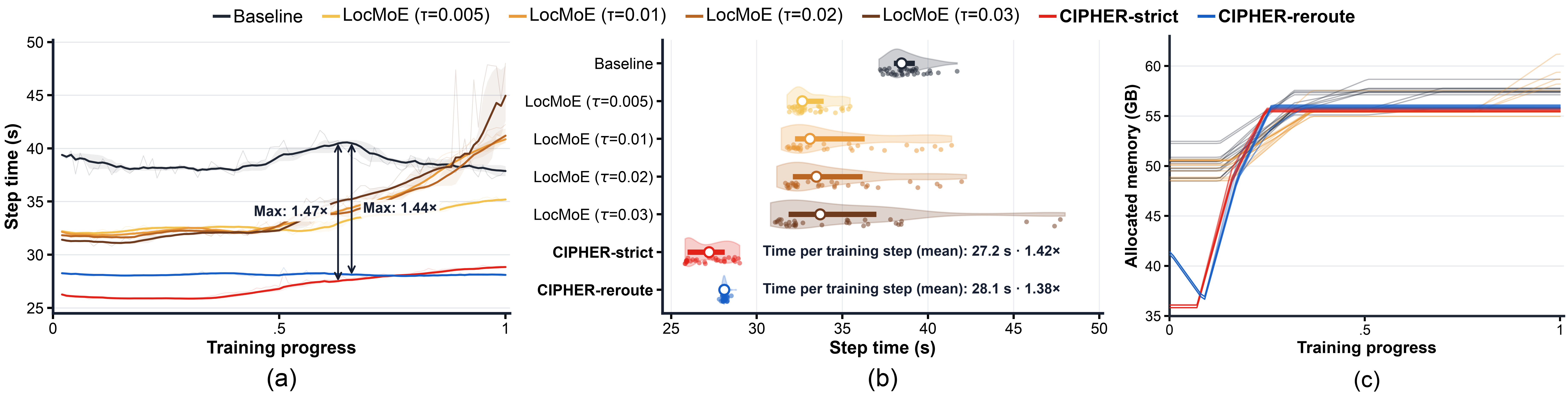}
    \caption{Training efficiency and memory footprint on DeepSeek-V4-Flash. (a) Per-step training time over the training progress, and CIPHER-MoE achieves up to 1.47$\times$ speedup. (b) Step-time distributions, where CIPHER-strict and CIPHER-reroute achieve an average speedup of 1.38–1.42$\times$ over the baseline. (c) Allocated memory throughout training.}
    \label{fig:time_memory}
\end{figure*}

\begin{figure*}[!t]
    \centering
    \vspace{-15pt}
    \includegraphics[width=0.85\linewidth]{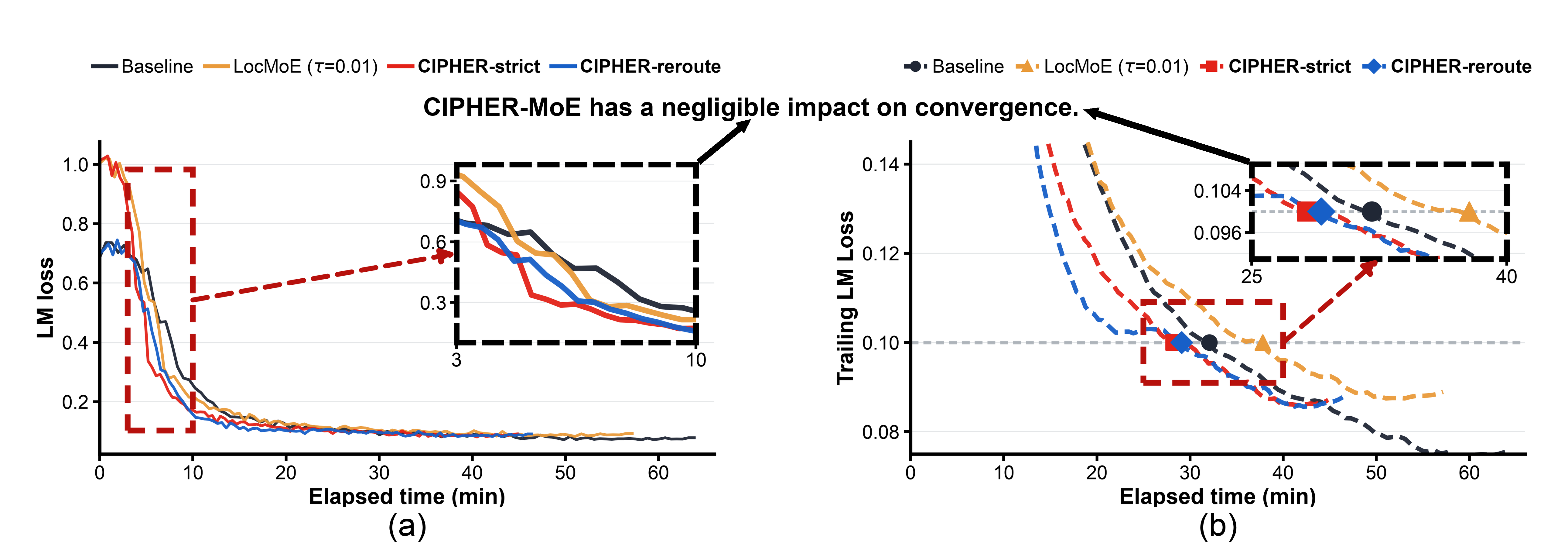}
    \caption{Training stability and convergence. (a) LM loss over elapsed training time. (b) Ten-step trailing LM loss and time to reach the 0.10 target.}
    \label{fig:lm_loss}
\end{figure*}

Training efficiency and accelerator memory usage are reported in Figure~\ref{fig:time_memory}. Since CIPHER-MoE targets very large-scale MoE models and few existing methods provide directly comparable results at this scale, we compare it with LocMoE~\citep{10.24963/ijcai.2024/705}, which has already been applied to large-cluster training. All reported training times are measured from end-to-end training runs. CIPHER-MoE achieves 1.42$\times$ speedups compared to other baselines. In the meantime, CIPHER-MoE limits the allocated memory to approximately 87.5\% of the device memory capacity. These gains arise from capacity-aware expert admission, which reduces the number of executed token–expert assignments and the associated computation and communication. The computational overheads introduced by the strict and reroute policies account for only 1.57\% and 3.85\% of the total training time, respectively, allowing both policies to preserve most of the efficiency gains.

\subsection{LLM Training Quality with CIPHER-MoE}

Figure~\ref{fig:lm_loss} demonstrates that CIPHER-MoE accelerates convergence without compromising training stability. Both CIPHER-strict and CIPHER-reroute exhibit smooth loss trajectories and converge reliably to the same low-loss regime as the reference methods. The detailed information of gradient norm is also provided in Appendix~\ref{sec:grad_norm}.
With the smooth training process of CIPHER-MoE, the post-trained LLM  achieves the optimal performance compared to the baselines, as presented in Table~\ref{tab:eval-domain}. The proposed CIPHER-MoE achieves the best quality-efficiency tradeoff regardless of the model sizes and architectures. 


\begin{figure*}[!t]
    \centering
    \includegraphics[width=0.9\linewidth]{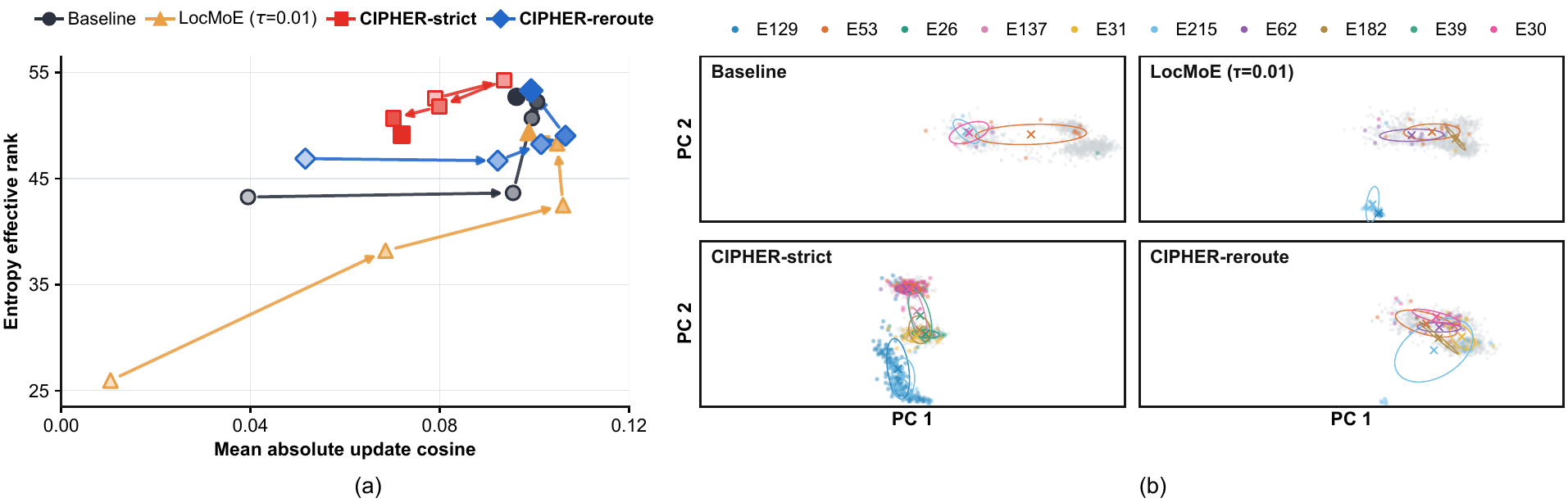}
    \caption{Router-update diversity and expert specialization. (a) Five consecutive training stages connected by arrows, with darker colors denoting later stages. Lower update-vector cosine similarity indicates clearer expert differentiation, while higher entropy effective rank indicates more dispersed, less homogeneous updates; the upper-left is better. (b) PCA of router-input hidden states. Compact same-color clusters indicate similar tokens routed to the same expert, while separation across colors indicates distinct token assignments and stronger specialization.}
    \label{fig:router_update}
\end{figure*}

\subsection{Expert Update Distributions and Degree of Homogenization}

As displayed in Figure~\ref{fig:router_update}, CIPHER-MoE consistently promotes stronger expert differentiation and avoids expert homogenization. CIPHER-strict exhibits the clearest separation in router adaptation. The agreement between the update dynamics and routed-representation geometry indicates that the resulting specialization reflects a persistent division of labor rather than a transient routing fluctuation. This behavior arises because affinity-aware capacity control suppresses weak or repetitive assignments, while selective rerouting directs overflow tokens toward compatible experts with available capacity, thereby preserving distinct expert learning signals.

\subsection{Training Efficiency and Evaluation of DeepSeek-V4-Pro}

\begin{table}[h!]
\centering
\begin{subtable}[t]{0.45\columnwidth}
\centering
\caption{Training efficiency comparison.}
\label{tab:training_efficiency}
\small
\setlength{\tabcolsep}{3pt}
\resizebox{\linewidth}{!}{%
\begin{tabular}{lcccc}
\toprule
\textbf{Method} & \textbf{GBS} & \shortstack{\textbf{Elapsed}\\\textbf{Time (s)}} & \textbf{Loss} & \shortstack{\textbf{Time}\\\textbf{Range (s)}} \\
\midrule
\shortstack[l]{DeepSeek-V4-Pro\\Baseline}
& 256
& \textemdash{} (OOM)
& \textemdash{}
& \textemdash{} \\
\multirow{2}{*}{\shortstack[l]{DeepSeek-V4-Pro\\CIPHER-MoE}}
& 256
& 45.16
& 0.29--0.43
& 5.50 \\
& 8192
& 538.65
& 1.41--1.51
& 91.11 \\
\bottomrule
\end{tabular}%
}
\end{subtable}%
\hfill
\begin{subtable}[t]{0.52\columnwidth}
\centering
\caption{Evaluation results of OR domain benchmarks (\%) for OR SFT.}
\label{tab:or_domain_evaluation}
\small
\setlength{\tabcolsep}{3pt}
\resizebox{\linewidth}{!}{%
\begin{tabular}{lccccc}
\toprule
\textbf{Method} & \textbf{NL4Opt} & \textbf{OptiBench} & \textbf{B4O-Feas.} & \textbf{B4O-OR} & \shortstack{\textbf{Weighted}\\\textbf{Avg.}} \\
\midrule
\shortstack[l]{DeepSeek-V4-Pro\\Fixed Router} & 89.27 & 64.67 & 74.42 & 54.31 & 68.59 \\
\shortstack[l]{DeepSeek-V4-Pro\\CIPHER-MoE} & 92.04 & 65.33 & 67.44 & 59.14 & 69.02 \\
\bottomrule
\end{tabular}%
}
\end{subtable}
\caption{DeepSeek-V4-Pro training efficiency and OR-domain evaluation.}
\label{tab:dsv4pro_efficiency_and_or_eval}
\end{table}

As shown in Table~\ref{tab:dsv4pro_efficiency_and_or_eval}, the DeepSeek-V4-Pro baseline fails to train under limited memory resources due to OOM errors, whereas CIPHER-MoE's lower memory footprint enables successful training without fixed routing. Training remains stable even at a global batch size of 8192. CIPHER-MoE also outperforms the fixed-router baseline on multiple OR benchmarks and achieves a higher weighted average score.

\section{Conclusion}
This work addresses the expert workload imbalance that limits the efficiency and stability of MoE training, with empirical evaluations spanning hundred-billion- to trillion-parameter models. 
We propose \texttt{CIPHER-MoE}, a plug-and-play approach that balances workloads by constraining overloaded experts through capacity-aware token selection and rerouting. 
More importantly, this work reveals the bounded relationship between expert capacity and workload distribution.
Experiments are conducted on the mainstream mega-sized MoE models ranging from 284B to 1.6T parameters, including DeepSeek-V3, DeepSeek-V4-Flash, GLM-5, and DeepSeek-V4-Pro. 
Across different architectures and datasets, CIPHER-MoE achieves measured training speedups of $1.10\times$--$1.94\times$, accelerating convergence while consistently preserving both domain-specific and general capabilities. 
These results demonstrate that CIPHER-MoE provides a simple, efficient, stable, and scalable foundation for large-scale MoE training.

\bibliography{iclr2027_conference}
\bibliographystyle{iclr2027_conference}

\appendix
\newpage
\section{Appendix}

\subsection{Gradient Dynamics and Training Stability of MoE Models}
\label{sec:grad_norm}
\begin{figure*}[h!]
    \centering
    \includegraphics[width=\linewidth]{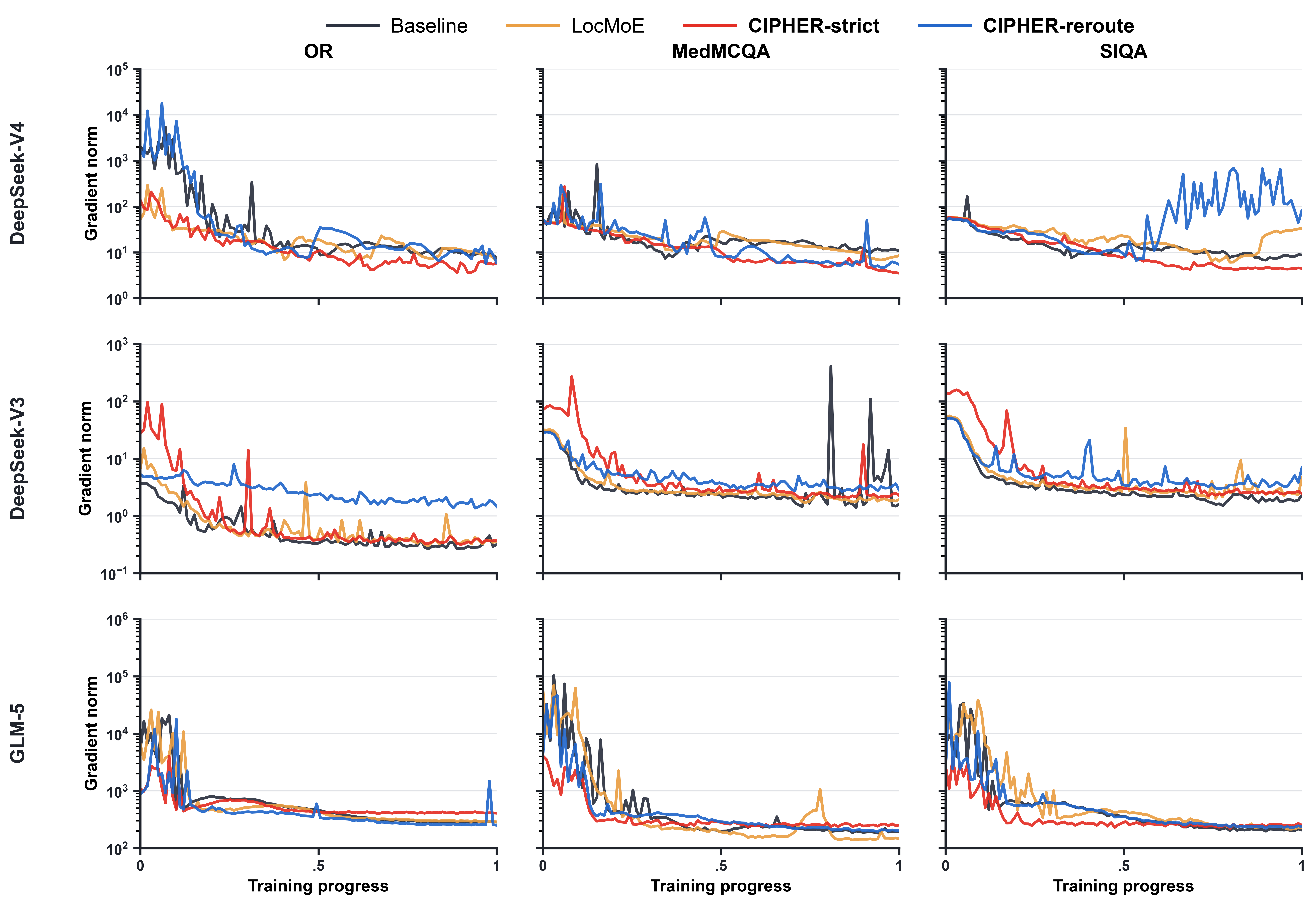}
    \caption{Gradient stability of CIPHER-MoE over training progress for DeepSeek-V4-Flash, DeepSeek-V3, and GLM-5 on MedMCQA, SIQA, and OR, shown on logarithmic scales. }
    \label{fig:grad_norm}
\end{figure*}

In Figure~\ref{fig:grad_norm}, across all evaluated model-dataset configurations, CIPHER-MoE maintains finite, non-degenerate gradient norms that generally decay from their initial transients into stable ranges, with no evidence of gradient explosion or vanishing. CIPHER-strict often produces smoother and lower gradient norms than the baseline, while CIPHER-reroute remains bounded despite occasional task-dependent fluctuations. Together with the consistently decreasing loss curves, these results indicate that CIPHER-MoE preserves well-behaved optimization dynamics and can provide greater training stability than the baseline. Overall, capacity-aware routing enables favorable convergence without destabilizing gradient-based training.

\begin{figure*}[t!]
    \centering
    \includegraphics[width=0.85\linewidth]{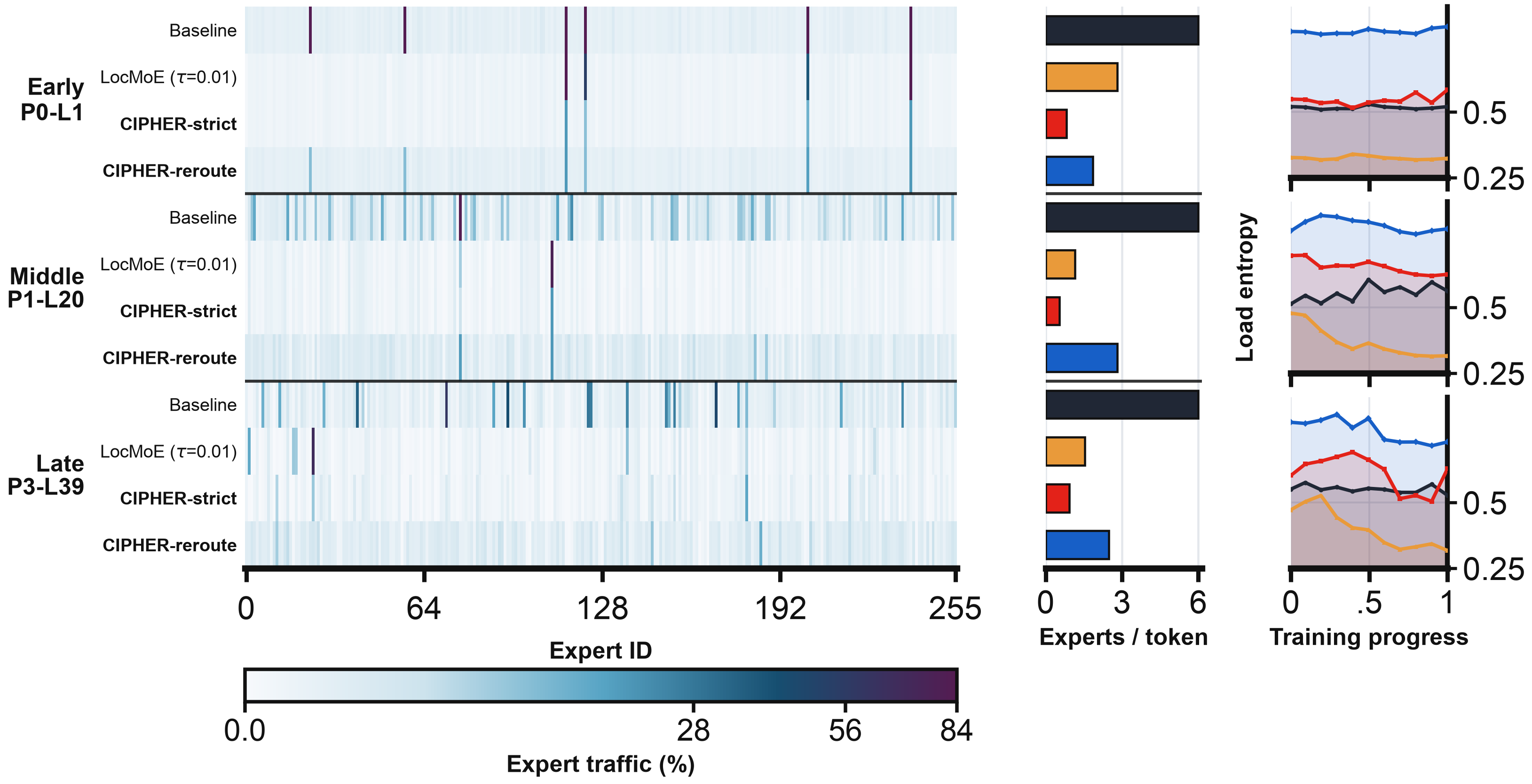}
    \caption{Expert routing behavior. Expert traffic across representative MoE layers (left), activated experts per token (middle), and load entropy during training (right). }
    \label{fig:routing_traffic}
\end{figure*}

In Figure~\ref{fig:routing_traffic}, across early, middle, and late layers, both CIPHER policies suppress the severe routing hotspots observed under the baseline and LocMoE, reducing peak expert traffic from as high as 83.6\% to 18.7\%; CIPHER-strict minimizes routing volume, whereas CIPHER-reroute achieves the most balanced expert utilization while retaining broader expert coverage. 

\subsection{Training Efficiency and Stability of Qwen Models}
\label{sec:qwen}
\begin{figure*}[h!]
    \centering
    \includegraphics[width=\linewidth]{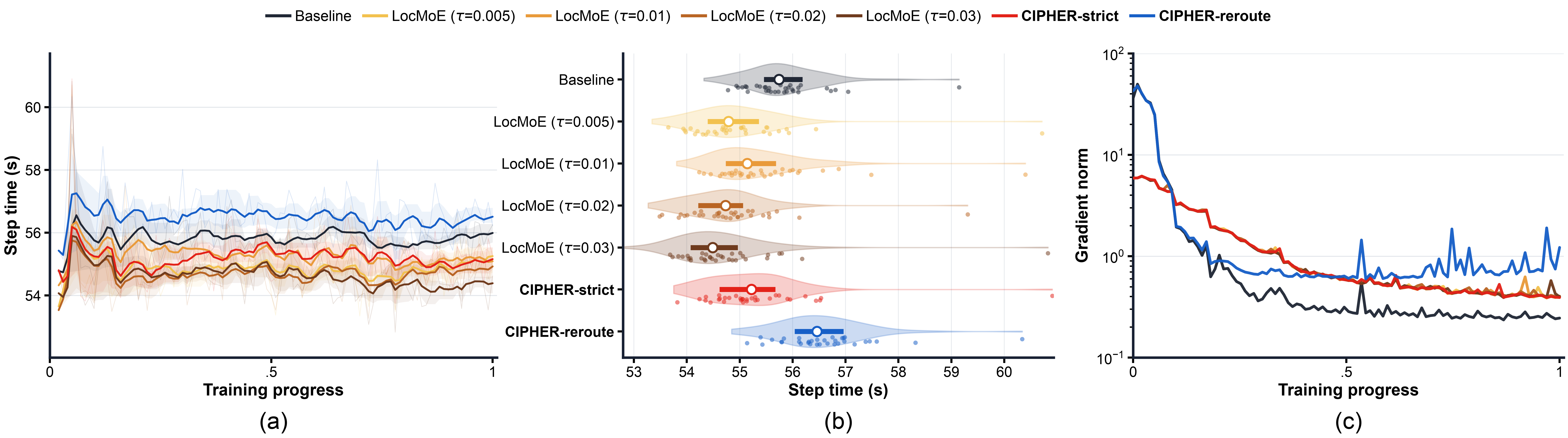}
    \caption{Training efficiency and optimization stability on Qwen3.6-35B-A3B. (a) Step time over training progress, shown with a 10-step rolling mean and interquartile range. (b) Distribution of post-warm-up step times. (c) Raw gradient norms on a logarithmic scale. CIPHER maintains stable step times and convergent gradient norms, with CIPHER-strict exhibiting smoother late-stage optimization.}
    \label{fig:qwen_eff_grad}
\end{figure*}

Qwen3.6-35B-A3B adopts hidden-state dimensions of 2048 for Qwen3.6-35B-A3B, whose relatively limited widths are consistent with failure to satisfy the favorable dimension-margin regime required by CIPHER-MoE in Table~\ref{tab:cipher_dimension}, as shown in Figure~\ref{fig:qwen_eff_grad}. Consequently, the observed training-efficiency improvement remains below 5\%, falling short of the gains achieved on larger MoE models. This contrast suggests that CIPHER-MoE is better suited to large-scale MoE architectures with higher-dimensional representations and may deliver increasingly substantial benefits as future models approach the 10-trillion-parameter scale.

\subsection{Evaluation results of general benchmarks}
\label{sec:general_benchmark}
\begin{table*}[h!]
\centering
\caption{Evaluation results of general benchmarks (\%) for OR SFT. }
\label{tab:eval-general-or}
\scriptsize
\setlength{\tabcolsep}{2.4pt}
\renewcommand{\arraystretch}{1.08}
\resizebox{\textwidth}{!}{%
\begin{tabular}{llrrrrrrrrrrr}
\toprule
\textbf{Model} & \textbf{Method} & \textbf{AIME-26} & \textbf{AMC} & \textbf{ARC} & \textbf{BBH} & \textbf{DROP} & \textbf{GPQA-D} & \textbf{GSM8K} & \textbf{HellaSwag} & \textbf{IFEval} & \textbf{MATH-500} & \textbf{MMLU-Pro} \\
\midrule
\textbf{DeepSeek-V4-Flash} & Baseline & 33.33 & \underline{71.60} & \underline{98.10} & \textbf{84.00} & 86.00 & 64.14 & 95.90 & \textbf{85.40} & \textbf{87.80} & 91.20 & 80.02 \\
 & \textbf{CIPHER-strict} & 30.00 & 70.90 & \textbf{98.20} & 71.20 & 84.50 & \underline{67.17} & 95.70 & 80.80 & 83.70 & 91.80 & 80.98 \\
 & \textbf{CIPHER-reroute} & \underline{40.00} & 70.60 & 97.80 & 58.00 & 72.30 & 55.56 & 95.70 & 81.80 & 80.40 & 90.40 & 75.47 \\
 & LocMoE ($\tau$=0.005) & 33.33 & 70.90 & \underline{98.10} & 71.00 & 86.30 & 66.67 & \textbf{96.80} & \underline{82.00} & \underline{86.50} & 92.20 & 80.59 \\
 & LocMoE ($\tau$=0.01) & 33.33 & 70.90 & \underline{98.10} & 71.30 & 86.30 & 65.15 & 96.30 & 81.80 & 85.20 & \underline{93.20} & 80.86 \\
 & LocMoE ($\tau$=0.02) & \textbf{46.67} & \underline{71.60} & 98.00 & 71.20 & \underline{86.40} & 65.15 & \underline{96.40} & 81.70 & 85.40 & 91.80 & \underline{81.44} \\
 & LocMoE ($\tau$=0.03) & 36.67 & \textbf{72.40} & \underline{98.10} & \underline{71.90} & \textbf{86.90} & \textbf{70.20} & 96.30 & 81.30 & 83.90 & \textbf{94.20} & \textbf{81.70} \\
\midrule
\textbf{DeepSeek-V3} & Baseline & \textbf{46.67} & \textbf{71.64} & \textbf{97.94} & \underline{87.87} & 87.19 & 52.53 & 95.68 & 75.97 & 73.74 & \textbf{88.80} & 75.51 \\
 & \textbf{CIPHER-strict} & \textbf{46.67} & 64.18 & \underline{97.91} & 85.99 & \textbf{89.24} & \textbf{57.58} & \underline{96.13} & \underline{87.45} & \underline{84.66} & 87.60 & \textbf{76.77} \\
 & \textbf{CIPHER-reroute} & \underline{43.33} & 62.69 & \underline{97.91} & 86.44 & \underline{88.51} & 52.53 & \textbf{96.29} & \textbf{87.48} & \textbf{93.05} & 88.00 & \underline{76.14} \\
 & LocMoE ($\tau$=0.01) & 40.00 & \underline{66.42} & 97.07 & \textbf{88.44} & 87.27 & \underline{56.57} & 95.68 & 80.65 & 81.65 & \underline{88.60} & 75.57 \\
\midrule
\textbf{GLM-5} & Baseline & \underline{63.33} & \textbf{86.57} & 98.20 & 93.20 & \underline{82.19} & 61.11 & 96.74 & 87.55 & 90.94 & \textbf{95.80} & \textbf{81.67} \\
 & \textbf{CIPHER-strict} & 53.33 & 81.34 & \textbf{98.59} & \underline{93.44} & 81.56 & \textbf{64.65} & \underline{97.19} & \textbf{88.19} & 91.31 & 92.80 & 80.15 \\
 & \textbf{CIPHER-reroute} & \textbf{66.67} & \underline{82.84} & \underline{98.37} & 91.34 & \textbf{82.96} & 58.59 & \textbf{97.80} & \underline{87.65} & \textbf{92.05} & \underline{95.20} & \underline{80.33} \\
 & LocMoE ($\tau$=0.01) & 60.00 & \underline{82.84} & 97.27 & \textbf{93.87} & 80.53 & \underline{63.64} & 96.21 & 85.22 & \underline{91.50} & \underline{95.20} & 77.85 \\
\bottomrule
\end{tabular}%
}
\end{table*}

\begin{table*}[h!]
\centering
\caption{Evaluation results of general benchmarks (\%) for MedMCQA SFT.}
\label{tab:eval-general-medmcqa}
\scriptsize
\setlength{\tabcolsep}{2.4pt}
\renewcommand{\arraystretch}{1.08}
\resizebox{\textwidth}{!}{%
\begin{tabular}{llrrrrrrrrrrr}
\toprule
\textbf{Model} & \textbf{Method} & \textbf{AIME-26} & \textbf{AMC} & \textbf{ARC} & \textbf{BBH} & \textbf{DROP} & \textbf{GPQA-D} & \textbf{GSM8K} & \textbf{HellaSwag} & \textbf{IFEval} & \textbf{MATH-500} & \textbf{MMLU-Pro} \\
\midrule
\textbf{DeepSeek-V4-Flash} & Baseline & \underline{50.00} & \textbf{80.60} & 97.50 & 85.00 & \textbf{85.20} & \underline{62.63} & 95.70 & \underline{80.20} & 70.20 & 92.60 & 76.38 \\
 & \textbf{CIPHER-strict} & \textbf{56.67} & 75.56 & \textbf{98.03} & \underline{85.16} & \underline{81.62} & \textbf{66.67} & \textbf{96.44} & 76.35 & \textbf{78.74} & \underline{94.00} & \underline{80.31} \\
 & \textbf{CIPHER-reroute} & \underline{50.00} & 75.56 & 97.74 & \textbf{86.13} & 81.40 & 60.10 & 96.13 & \textbf{80.33} & 74.12 & 92.80 & 78.77 \\
 & LocMoE ($\tau$=0.01) & \textbf{56.67} & \underline{80.00} & \underline{97.86} & 84.53 & 81.51 & \textbf{66.67} & \underline{96.29} & 76.86 & \underline{77.45} & \textbf{94.20} & \textbf{80.38} \\
\midrule
\textbf{DeepSeek-V3} & Baseline & \underline{46.67} & \textbf{77.61} & 96.65 & \underline{87.70} & 82.65 & \textbf{60.61} & \textbf{96.66} & 74.38 & 80.10 & \underline{92.20} & 74.28 \\
 & \textbf{CIPHER-strict} & \underline{46.67} & 74.63 & \textbf{97.97} & 86.28 & \textbf{86.77} & 55.56 & \underline{96.29} & \textbf{87.48} & \underline{82.07} & 88.00 & \textbf{76.32} \\
 & \textbf{CIPHER-reroute} & \textbf{53.33} & 76.12 & \underline{97.86} & 85.33 & 85.66 & \underline{59.09} & 95.83 & \underline{80.87} & 80.41 & 90.20 & 74.75 \\
 & LocMoE ($\tau$=0.01) & 43.33 & \underline{76.86} & 96.99 & \textbf{88.33} & \underline{85.89} & 57.07 & \textbf{96.66} & 75.08 & \textbf{83.93} & \textbf{92.80} & \underline{75.01} \\
\midrule
\textbf{GLM-5} & Baseline & \textbf{63.33} & \underline{86.57} & 98.34 & 91.49 & \textbf{84.49} & 61.11 & 96.82 & \textbf{87.46} & 90.76 & 94.80 & 80.38 \\
 & \textbf{CIPHER-strict} & \underline{60.00} & 85.07 & \textbf{98.68} & \underline{92.58} & 82.01 & \underline{62.63} & \textbf{97.88} & 86.25 & 90.20 & 95.40 & \underline{81.38} \\
 & \textbf{CIPHER-reroute} & \underline{60.00} & \textbf{88.06} & \underline{98.39} & 92.40 & 79.38 & \textbf{65.15} & 96.89 & 85.74 & \textbf{94.64} & \underline{95.80} & \textbf{81.82} \\
 & LocMoE ($\tau$=0.01) & \underline{60.00} & 82.09 & 98.00 & \textbf{93.12} & \underline{82.50} & \textbf{65.15} & \underline{96.97} & \underline{86.31} & \underline{91.31} & \textbf{96.20} & 75.07 \\
\bottomrule
\end{tabular}%
}
\end{table*}

\begin{table*}[h!]
\centering
\caption{Evaluation results of general benchmarks (\%) for SIQA SFT.}
\label{tab:eval-general-siqa}
\scriptsize
\setlength{\tabcolsep}{2.4pt}
\renewcommand{\arraystretch}{1.08}
\resizebox{\textwidth}{!}{%
\begin{tabular}{llrrrrrrrrrrr}
\toprule
\textbf{Model} & \textbf{Method} & \textbf{AIME-26} & \textbf{AMC} & \textbf{ARC} & \textbf{BBH} & \textbf{DROP} & \textbf{GPQA-D} & \textbf{GSM8K} & \textbf{HellaSwag} & \textbf{IFEval} & \textbf{MATH-500} & \textbf{MMLU-Pro} \\
\midrule
\textbf{DeepSeek-V4-Flash} & Baseline & \underline{53.33} & \underline{75.40} & 96.30 & \textbf{87.90} & \textbf{84.70} & \underline{60.10} & 96.10 & \underline{87.10} & 79.30 & 90.80 & 77.68 \\
 & \textbf{CIPHER-strict} & 50.00 & \textbf{77.78} & \underline{98.03} & 85.64 & 81.53 & \textbf{62.63} & \underline{96.44} & 85.12 & 80.59 & \textbf{92.00} & \underline{78.26} \\
 & \textbf{CIPHER-reroute} & \textbf{60.00} & 73.33 & \textbf{98.17} & \underline{85.73} & 79.46 & \underline{60.10} & 95.75 & \textbf{88.06} & \textbf{81.48} & \underline{91.60} & 74.67 \\
 & LocMoE ($\tau$=0.01) & \underline{53.33} & 73.33 & 97.91 & \underline{85.73} & \underline{81.91} & 56.57 & \textbf{96.51} & 84.87 & \underline{81.33} & 91.20 & \textbf{78.27} \\
\midrule
\textbf{DeepSeek-V3} & Baseline & \underline{53.33} & \underline{72.39} & 96.62 & \underline{85.30} & 84.31 & 54.55 & 93.10 & 78.87 & 76.14 & 87.60 & 71.67 \\
 & \textbf{CIPHER-strict} & 46.67 & 65.67 & \textbf{97.94} & \textbf{86.35} & \textbf{88.39} & \textbf{57.58} & \textbf{95.98} & \textbf{87.61} & \textbf{92.81} & 87.40 & \textbf{73.99} \\
 & \textbf{CIPHER-reroute} & \underline{53.33} & 68.66 & \underline{97.86} & 84.23 & \underline{87.45} & \underline{56.57} & 95.00 & \underline{85.23} & \underline{89.09} & \underline{88.60} & 71.01 \\
 & LocMoE ($\tau$=0.01) & \textbf{56.67} & \textbf{73.88} & 97.10 & 81.97 & 87.21 & 54.55 & \underline{95.68} & 80.01 & 87.29 & \textbf{90.80} & \underline{72.22} \\
\midrule
\textbf{GLM-5} & Baseline & \underline{53.33} & 82.84 & 97.49 & 91.49 & \underline{83.66} & \underline{61.11} & 95.91 & 86.48 & \underline{91.50} & \underline{93.80} & 80.56 \\
 & \textbf{CIPHER-strict} & \textbf{56.67} & \underline{84.33} & \textbf{98.03} & \textbf{93.56} & 83.47 & \textbf{62.63} & \textbf{97.04} & \textbf{89.12} & \textbf{91.68} & 92.40 & \underline{81.22} \\
 & \textbf{CIPHER-reroute} & \textbf{56.67} & \textbf{85.07} & \underline{97.89} & \underline{92.87} & \textbf{84.55} & \underline{61.11} & \underline{96.89} & \underline{88.74} & 90.57 & \underline{93.80} & \textbf{82.68} \\
 & LocMoE ($\tau$=0.01) & 50.00 & 82.09 & 97.66 & 92.23 & 81.75 & \textbf{62.63} & 95.83 & 86.88 & 90.20 & \textbf{95.20} & 75.86 \\
\bottomrule
\end{tabular}%
}
\end{table*}

General capabilities are evaluated across five categories: mathematics (AIME-26~\citep{maa2026aime1,maa2026aime2}, AMC~\citep{maa2024amc12}, GSM8K~\citep{cobbe2021training}, and MATH-500~\citep{lightman2024let}), general knowledge (ARC~\citep{clark2018think}, GPQA-D~\citep{rein2023gpqa}, MMLU-Pro~\citep{wang2024mmlu}, and HellaSwag~\citep{zellers2019hellaswag}), logical reasoning (BBH~\citep{suzgun2023challenging}), reading comprehension (DROP~\citep{dua2019drop}), and instruction following (IFEval~\citep{zhou2023instruction}). Tables~\ref{tab:eval-general-or}, \ref{tab:eval-general-medmcqa}, and \ref{tab:eval-general-siqa} demonstrate that CIPHER-MoE can strengthen broad transfer performance rather than merely preserve it. For DeepSeek-V3 after SIQA SFT, CIPHER-strict improves the macro-average across ARC, BBH, GPQA-D, HellaSwag, and MMLU-Pro by 4.3\% relative to the baseline. This pattern suggests that selective routing suppresses low-value signals and concentrates expert updates on more informative token--expert associations. The efficiency benefits of CIPHER-MoE therefore need not come at the expense of general capability and may instead be accompanied by better transferable reasoning.

\subsection{Capacity analysis of Theorem~\ref{thm:cipher_capacity}}
\label{app:cipher_capacity}


\subsubsection{Comparison task and assumptions}
\label{sec:cipher_assumptions}

Consider expert $e$ and its candidate pool in one micro-batch. Let $N_e\geq1$ be the number of offered tokens, and let $C\in\{1,\ldots,N_e\}$ be the number retained for expert execution. A token is \emph{discriminative} if it carries the expert-specific signal specified by the model below. Write $D_i$ for its indicator and $A_e=\{\sum_{i=1}^{N_e}D_i\geq1\}$ for the event that the offered pool contains such a token. The common retention target is given in equation~\ref{eq:cipher_retention_target}.
Natural-order admission executes the first $C$ offered tokens, whereas CIPHER admission ranks the available candidates before allocating its $C$ slots. Computing affinities and selecting candidates are not counted as expert-execution capacity. The comparison therefore concerns signal retention per execution slot; its implications for end-to-end training time are evaluated separately.

\paragraph{A1: Exposure, geometry, and signal availability.}
The offered tokens are independent draws from a common exposure distribution, with discriminative probability $\pi_e\in(0,1)$ and
\begin{equation}
    \Pr(A_e)=1-(1-\pi_e)^{N_e}\geq a_0>0.
    \label{eq:cipher_availability}
\end{equation}
For the geometric reference model, let $X\sim\operatorname{Unif}(S^{d-1})$, let $\vu_e$ be the unit gating direction, and let $\mathcal R_e$ denote exposure to expert $e$. Suppose the discriminative event $\mathcal D_e$ and exposure probability satisfy
\begin{equation}
    \mathcal D_e\subseteq\{\langle\vu_e,X\rangle\geq\delta\},
    \qquad \Pr(\mathcal R_e)\geq\frac{1}{\beta E},
    \qquad 0<\delta<1,\quad\beta\geq1.
    \label{eq:cipher_exposure}
\end{equation}
The offered distribution is the exposure-conditioned distribution, rather than an unconditioned uniform spherical distribution. Consequently,
\begin{equation}
    \pi_e=\Pr(\mathcal D_e\mid\mathcal R_e)
    \leq\beta E\Pr(\mathcal D_e)
    \leq\beta E p_\delta(d),
    \qquad
    p_\delta(d)=\Pr(|\langle\vu_e,X\rangle|\geq\delta).
    \label{eq:cipher_exposure_probability}
\end{equation}
No independence between discrimination and exposure is required for this inequality. The two-sided spherical probability gives a conservative bound for the one-sided discriminative event. Equal-norm orthogonal gates under the symmetric reference model yield $\beta=1$; learned gates, expert biases, and Top-$K$ selection require their own exposure justification. The margin $\delta$ is a certified property of the signal model and is distinct from the algorithmic affinity threshold $\tau$.

\paragraph{A2: Candidate and threshold recall.}
Conditional on $A_e$, choose one discriminative token as a \emph{witness} according to a fixed rule. Let $\mathcal V_e$ denote the event that this witness reaches the expert's ranking pool and survives thresholding. Assume
\begin{equation}
    \Pr(\mathcal V_e\mid A_e)\geq1-\eta,
    \qquad 0\leq\eta<\varepsilon.
    \label{eq:cipher_recall}
\end{equation}
This explicitly accounts for signal loss before capacity truncation. Increasing execution capacity cannot recover a witness that has already been excluded from the ranking pool.

\paragraph{A3: Stable conditional ranking.}
Condition on $A_e$ and $\mathcal V_e$. An expert ranks candidates using affinity and a fixed deterministic rule for ties. Expose the competing candidates sequentially, and let $Y_j$ indicate whether competitor $j$ precedes the witness. For the information $\mathcal F_{j-1}$ revealed by previous competitors, assume deterministic bounds
\begin{equation}
    \Pr(Y_j=1\mid\mathcal F_{j-1},A_e,\mathcal V_e)\leq q_j,
    \qquad 0\leq q_j\leq1,
    \qquad \sum_{j=1}^{N_e-1}q_j\leq\mu.
    \label{eq:cipher_blocker_assumption}
\end{equation}
Here, $\mu$ bounds the total conditional number of competitors that can rank ahead of the witness. If fewer than $N_e$ candidates reach the ranking pool, the remaining indicators are set to zero. With $S=\sum_jY_j$, the witness occupies position $S+1$. This assumption allows dependence between candidates, provided the bound holds after witness selection and after observing earlier competitors. A marginal ranking-error rate alone does not imply this conditional guarantee.

\subsubsection{Capacity separation for affinity-based admission}
\label{sec:cipher_capacity_theorem}

We prove the capacity bounds in Theorem~\ref{thm:cipher_capacity}, using $B_e$ and $G(d,\delta)$ as defined in equation~\ref{eq:cipher_sufficient_budget}. For any prescribed reduction factor $\kappa>1$, we also derive the following sufficient condition for $C^*_{\mathrm{CIPHER}}\leq C^*_{\mathrm{natural}}/\kappa$:
\begin{equation}
    d\geq1+
    \frac{2\log\!\left(\kappa\beta E B_e/[(1-\varepsilon)a_0]\right)}
    {-\log(1-\delta^2)}.
    \label{eq:cipher_dimension_threshold}
\end{equation}

Theorem~\ref{thm:cipher_capacity} identifies two different controls on execution capacity. Natural-order admission must allocate enough slots to encounter a scarce discriminative token. Affinity admission instead needs enough slots to accommodate candidates ranked ahead of an available witness. When this conditional ranking budget is small, the latter policy can operate with substantially fewer slots. The quantity $B_e$ is a sufficient budget, rather than an exact minimum or an optimal training-time configuration.

\paragraph{Spherical concentration.}
We first establish the finite-dimensional geometric bound used in the theorem. Rotational symmetry and the normalized Gaussian representation of a spherical vector give
\begin{equation}
    p_\delta(d)
    =1-I_{\delta^2}\!\left(\frac12,\frac{d-1}{2}\right)
    \leq(1-\delta^2)^{(d-1)/2}
    =e^{-G(d,\delta)},
    \label{eq:cipher_spherical_cap}
\end{equation}
where $I_z(a,b)$ is the regularized incomplete beta function. To prove the inequality, let $Z_1,\ldots,Z_d$ be independent standard Gaussian variables, set $W=\sum_{j=2}^dZ_j^2$, and write $X=Z/\|Z\|$. Then $X_1^2=Z_1^2/(Z_1^2+W)$ has the beta distribution appearing above. With $b=\delta^2/(1-\delta^2)$, the cap event is $Z_1^2\geq bW$. The Gaussian tail satisfies $2[1-\Phi(v)]\leq e^{-v^2/2}$ for $v\geq0$, where $\Phi$ is the standard Gaussian cumulative distribution function. Indeed, the difference between the right and left sides vanishes at zero and at infinity, and its derivative is $e^{-v^2/2}(\sqrt{2/\pi}-v)$, so it remains nonnegative. Conditioning on $W$ therefore yields
\begin{equation}
\begin{aligned}
    p_\delta(d)
    &=\mathbb E\!\left[2\{1-\Phi(\sqrt{bW})\}\right]
    \leq\mathbb E e^{-bW/2}\\
    &=\prod_{j=2}^d\mathbb E e^{-bZ_j^2/2}
    =(1+b)^{-(d-1)/2}
    =(1-\delta^2)^{(d-1)/2}.
\end{aligned}
\label{eq:cipher_cap_proof}
\end{equation}
This bound holds at finite $d$ and does not require an asymptotic approximation. In particular, $G(d,\delta)\geq(d-1)\delta^2/2$.

\paragraph{Necessary capacity for natural-order admission.}
Success within the first $C$ slots implies $A_e$. Independence in A1 thus gives
\begin{equation}
\begin{aligned}
    P_{\mathrm{natural}}(C\mid A_e)
    &=\frac{1-(1-\pi_e)^C}{1-(1-\pi_e)^{N_e}},\\
    C^*_{\mathrm{natural}}
    &=\left\lceil
    \frac{\log\!\left(1-(1-\varepsilon)[1-(1-\pi_e)^{N_e}]\right)}
    {\log(1-\pi_e)}\right\rceil.
\end{aligned}
    \label{eq:cipher_natural_exact}
\end{equation}
The exact expression follows by solving $P_{\mathrm{natural}}(C\mid A_e)\geq1-\varepsilon$; division by $\log(1-\pi_e)<0$ reverses the inequality. For a simpler necessary bound, $1-(1-\pi_e)^C\leq C\pi_e$ and equation~\ref{eq:cipher_availability} imply
\begin{equation}
    C^*_{\mathrm{natural}}
    \geq\frac{(1-\varepsilon)a_0}{\pi_e}
    \geq\frac{(1-\varepsilon)a_0}{\beta E}e^{G(d,\delta)},
    \label{eq:cipher_natural_lower}
\end{equation}
where the second inequality uses equations~\ref{eq:cipher_exposure_probability} and~\ref{eq:cipher_spherical_cap}. The union-bound argument requires only the marginal scarcity bound and positive availability; independence is needed for the exact expression in equation~\ref{eq:cipher_natural_exact}.

\paragraph{Conditional concentration of competing candidates.}
Write $\mathbb Q$ for the probability law conditioned on $A_e$ and $\mathcal V_e$. For $\theta>0$, A3 gives
\begin{equation}
\begin{aligned}
    \mathbb E_{\mathbb Q}[e^{\theta Y_j}\mid\mathcal F_{j-1}]
    &=1+(e^\theta-1)\mathbb E_{\mathbb Q}[Y_j\mid\mathcal F_{j-1}]\\
    &\leq1+q_j(e^\theta-1)
    \leq\exp\{q_j(e^\theta-1)\}.
\end{aligned}
\label{eq:cipher_conditional_mgf}
\end{equation}
Iterating conditional expectations and using $\sum_jq_j\leq\mu$ yields $\mathbb E_{\mathbb Q}e^{\theta S}\leq\exp\{\mu(e^\theta-1)\}$. Exponential Markov and minimization over $\theta>0$ then give, for $\mu>0$ and $x>0$,
\begin{equation}
\begin{aligned}
    \Pr_{\mathbb Q}(S\geq\mu+x)
    &\leq\exp\{-\mu h(x/\mu)\}
    \leq\exp\!\left\{-\frac{x^2}{2(\mu+x/3)}\right\},\\
    h(v)&=(1+v)\log(1+v)-v.
\end{aligned}
    \label{eq:cipher_blocker_tail}
\end{equation}
The first exponent follows by setting $\theta=\log(1+x/\mu)$. For the second, the series inequality $e^\theta-1-\theta\leq\theta^2/[2(1-\theta/3)]$, valid for $0<\theta<3$, follows from $k!\geq2\cdot3^{k-2}$ for $k\geq2$. Substituting $\theta=x/(\mu+x/3)$ into the Markov exponent gives the stated bound. If $\mu=0$, every $q_j$ is zero, so $S=0$ almost surely under $\mathbb Q$ and one slot suffices for an admitted witness.

\paragraph{Sufficient CIPHER capacity and proof of separation.}
Choose
\label{app:suff_proof}

\begin{equation}
    x=\frac{\ell}{3}+\sqrt{2\mu\ell+\frac{\ell^2}{9}},
    \qquad x^2=2\ell(\mu+x/3).
    \label{eq:cipher_tail_inversion}
\end{equation}

Equation~\ref{eq:cipher_blocker_tail} bounds the ranking failure by $e^{-\ell}=\varepsilon-\eta$. Since the witness has rank $S+1$, capacity $\lceil1+\mu+x\rceil$ is sufficient. If this exceeds $N_e$, admitting all available candidates eliminates capacity-induced failure, giving $B_e$ in equation~\ref{eq:cipher_sufficient_budget}. The same expression remains sufficient when $\mu=0$ or $N_e=1$, although it can be conservative. Accounting for candidate-recall failure gives
\begin{equation}
    \Pr(\text{failure}\mid A_e)
    \leq\eta+(1-\eta)(\varepsilon-\eta)
    \leq\varepsilon.
    \label{eq:cipher_total_failure}
\end{equation}
Thus $C^*_{\mathrm{CIPHER}}\leq B_e$. Together with equation~\ref{eq:cipher_natural_lower}, this proves equation~\ref{eq:cipher_capacity_separation}. Requiring the natural-order lower bound to be at least $\kappa B_e$ is equivalent to
\begin{equation}
    G(d,\delta)\geq
    \log\frac{\kappa\beta E B_e}{(1-\varepsilon)a_0},
    \label{eq:cipher_separation_condition}
\end{equation}
which gives equation~\ref{eq:cipher_dimension_threshold} after substituting the definition of $G$. This completes the proof of Theorem~\ref{thm:cipher_capacity}.

\paragraph{Finite-pool feasibility.}
Availability imposes the additional requirement
\begin{equation}
    N_e\geq\frac{\log(1-a_0)}{\log(1-\pi_e)}.
    \label{eq:cipher_pool_feasibility}
\end{equation}
Ranking cannot create a discriminative token absent from the offered pool. If the claimed natural-order lower bound exceeds $N_e$, the assumed scarcity and availability cannot both hold for that pool. The theorem does not prescribe execution of more tokens than are available. The degenerate case $\pi_e=0$ violates positive availability, whereas $\pi_e=1$ requires only one natural-order slot and admits no nontrivial separation of this kind.

The assumptions are jointly feasible. For example, consider a single expert, identity exposure, independent spherical tokens, and $\mathcal D_e=\{\langle\vu_e,X\rangle\geq\delta\}$. Choose a sufficiently large pool to maintain $a_0$, and choose its highest-projection discriminative token as the witness. Ranking by exact projection places no competitor ahead of this witness, up to probability-zero ties. With complete candidate recall, $\eta=\mu=0$ and one affinity-selected slot suffices, whereas natural-order admission still follows equation~\ref{eq:cipher_natural_exact}. This construction establishes compatibility of the assumptions without asserting that a trained checkpoint follows the reference distribution.

\paragraph{Retention guarantee for reroute.}
For the same input micro-batch, let $\mathcal T_e^{\mathrm S}$ be the tokens retained by expert $e$ under strict selection and $\mathcal T_e^{\mathrm R}$ its final token set after rerouting. If rerouting preserves all strictly accepted assignments and adds only tokens that satisfy the destination expert's similarity threshold within its remaining capacity, then
\begin{equation}
    \mathcal T_e^{\mathrm S}\subseteq\mathcal T_e^{\mathrm R},
    \qquad |\mathcal T_e^{\mathrm R}|\leq C_e,
    \label{eq:cipher_reroute_retention}
\end{equation}
Every discriminative witness retained by strict is therefore also retained by reroute. The strict-retention event is a subset of the reroute-retention event, so reroute inherits the same lower bound on the conditional retention probability under the assumptions of Theorem~\ref{thm:cipher_capacity}. This guarantee requires preserving accepted assignments and respecting each expert's capacity; it does not apply if rerouting evicts a strictly accepted token. Additional assignments can change the expert mixture and the resulting training updates, so the retention guarantee does not imply unchanged model outputs or improved downstream performance.

\subsubsection{Sharper capacity bounds and stability conditions}
\label{sec:cipher_stability}

\paragraph{Sharper conditional ranking bounds.}
When A3 admits a common bound $q_j\leq q$, retaining the exact Bernoulli moment-generating function gives
\begin{equation}
\begin{aligned}
    \mathbb E_{\mathbb Q}e^{\theta S}&\leq(1-q+qe^\theta)^{N_e-1},\\
    \Pr_{\mathbb Q}(S\geq k)
    &\leq\exp\!\left[-(N_e-1)\operatorname{kl}
    \left(\frac{k}{N_e-1}\middle\|q\right)\right],
\end{aligned}
    \label{eq:cipher_kl_tail}
\end{equation}
for $N_e\geq2$, $0<q<1$, and $(N_e-1)q<k\leq N_e-1$, where $\operatorname{kl}(u\|q)=u\log(u/q)+(1-u)\log[(1-u)/(1-q)]$. Optimizing the Markov exponent yields $e^\theta=u(1-q)/[q(1-u)]$ for $u=k/(N_e-1)<1$; the endpoint $u=1$ follows by taking $\theta\to\infty$. Numerically inverting this bound can reduce the sufficient integer budget. If the blockers are independent Bernoulli variables under $\mathbb Q$, the exact witness-retention budget is the corresponding binomial quantile,
\begin{equation}
    C^*_{\mathrm{rank}}(\alpha)
    =1+\min\{k\in\{0,\ldots,N_e-1\}:
    \Pr(\operatorname{Bin}(N_e-1,q)\leq k)\geq1-\alpha\}.
    \label{eq:cipher_binomial_capacity}
\end{equation}
Here $\alpha$ is the conditional ranking-failure allowance, and $\alpha=\varepsilon-\eta$ accounts for candidate recall as above. A simpler sufficient budget associated with the comparison is
\begin{equation}
    \widetilde C
    =\min\left\{N_e,
    \left\lceil\max\left\{2N_eq,\frac{16}{3}\log\frac1\alpha\right\}\right\rceil\right\}.
    \label{eq:cipher_etr_capacity}
\end{equation}
Indeed, with $m_0=(N_e-1)q\leq C/2$, the Bernstein exponent satisfies $3(C-m_0)^2/[2(C+2m_0)]\geq3C/16$. Admitting the entire pool handles budgets exceeding $N_e$; one slot suffices when $q=0$. These bounds concern the chosen witness, whereas retaining any one of several discriminative tokens may require fewer slots.

\paragraph{Data drift and affinity calibration.}
To connect A3 with changing training data, suppose a relevant signal has ideal normalized affinity at least $\delta_0$, and data drift reduces this margin by at most $\Delta$. Let $c^\star_{j,e}$ and $c_{j,e}$ be ideal and learned affinities, respectively. Assume both scores in a witness--competitor comparison have error at most $\zeta$, except with conditional probability $\varrho$. Also assume the conditional competitor tail satisfies
\begin{equation}
    \Pr(c^\star_{j,e}\geq z\mid\mathcal F_{j-1},A_e,\mathcal V_e)
    \leq\chi\Pr(\langle\vu_e,X\rangle\geq z)+\nu,
    \qquad \chi\geq1,\quad\nu\geq0.
    \label{eq:cipher_drift_envelope}
\end{equation}
The parameters $\chi$ and $\nu$ account for tail inflation and contamination after conditioning on exposure, witness admission, and previous competitors. Unconditional isotropy or calibration accuracy does not establish this bound.

On a calibrated comparison, a competitor preceding the witness must have ideal affinity at least
\begin{equation}
    c^\star_{j,e}\geq c_{j,e}-\zeta
    \geq c_{*,e}-\zeta
    \geq c^\star_{*,e}-2\zeta
    \geq\delta_0-\Delta-2\zeta
    \equiv\delta_{\mathrm{eff}}>0.
    \label{eq:cipher_effective_margin}
\end{equation}
For $\delta_{\mathrm{eff}}<1$, a union bound and the one-sided spherical tail yield
\begin{equation}
    q_{\max}\leq
    \min\left\{1,\varrho+\nu+
    \frac{\chi}{2}e^{-G(d,\delta_{\mathrm{eff}})}\right\},
    \qquad \mu=(N_e-1)q_{\max}.
    \label{eq:cipher_drift_blockers}
\end{equation}
The factor $1/2$ arises from the one-sided competitor event; equation~\ref{eq:cipher_exposure_probability} used a conservative two-sided bound. A bounded total ranking budget is obtained along configurations satisfying $N_e(\varrho+\nu)=O(1)$ and $N_e\chi e^{-G(d,\delta_{\mathrm{eff}})}=O(1)$. With a confidence gap $\varepsilon-\eta$ bounded away from zero, this gives $B_e=O(1)$. Thus, a larger candidate pool requires correspondingly stronger control of ranking errors.

For a certified budget $B_e$ and desired factor $\kappa$, write $L_\kappa=\log[\kappa\beta E B_e/((1-\varepsilon)a_0)]$. Applying equation~\ref{eq:cipher_separation_condition} with the conservative margin $\delta_{\mathrm{eff}}$ gives the sufficient drift tolerance
\begin{equation}
    \Delta+2\zeta\leq
    \delta_0-\sqrt{1-\exp\{-2L_\kappa/(d-1)\}}.
    \label{eq:cipher_drift_tolerance}
\end{equation}
The right side must be nonnegative, and this condition must hold jointly with candidate recall, availability, and the bound on $\mu$. If the data domain or router changes enough to invalidate these conditions, the admission budget or threshold must be reconsidered.

\paragraph{Dimension and finite-horizon scaling.}
Theorem~\ref{thm:cipher_capacity} gives
\begin{equation}
    \frac{C^*_{\mathrm{CIPHER}}}{C^*_{\mathrm{natural}}}
    \leq\frac{\beta E B_e}{(1-\varepsilon)a_0}e^{-G(d,\delta)}.
    \label{eq:cipher_capacity_ratio}
\end{equation}
A growing separation follows if $G(d,\delta)-\log(\beta E B_e/a_0)\to\infty$ with fixed confidence. For bounded $\mu$ and fixed exposure and availability parameters, $d\delta(d)^2\to\infty$ is sufficient. In contrast, when $\delta=c/\sqrt d$ for fixed $c>0$, the normalized Gaussian representation gives $\sqrt d X_1\Rightarrow\mathcal N(0,1)$ and hence $p_{c/\sqrt d}(d)\to2[1-\Phi(c)]$. The geometric exponent remains of constant order, so increasing dimension alone does not yield an increasing separation in this regime. These conclusions require availability to remain valid, potentially through a growing offered pool. Nominal hidden width or total parameter count does not by itself certify the effective spherical dimension of an anisotropic learned representation.

For $H$ expert--micro-batch pairs, set $\varepsilon=\varepsilon_{\mathrm{tot}}/H$, require $\eta<\varepsilon$, and recompute $B_e$. If the assumptions hold for each pair, a union bound gives
\begin{equation}
    \Pr\!\left(\bigcup_{h=1}^{H}(F_h\cap A_h)\right)
    \leq\sum_{h=1}^{H}\Pr(F_h\mid A_h)\Pr(A_h)
    \leq\varepsilon_{\mathrm{tot}},
    \label{eq:cipher_finite_horizon}
\end{equation}
where $F_h$ denotes retention failure and $A_h$ signal availability. Independence between micro-batches is unnecessary. The guarantee covers failures to retain an offered signal; it neither guarantees signal availability nor establishes an unlimited-horizon training guarantee.

\paragraph{Illustrative capacity regimes.}
For $E=256$, $\beta=1$, $\varepsilon=0.1$, $\eta=0.01$, $a_0=0.9$, and $\mu=1$, equation~\ref{eq:cipher_sufficient_budget} gives $B_e=6$ when $N_e\geq6$. Substituting these values into equation~\ref{eq:cipher_dimension_threshold} gives Table~\ref{tab:cipher_dimension}. These are conditional illustrations rather than fitted properties of the evaluated models. In particular, $N_e\geq6$ establishes only the displayed budget; the offered pool must also satisfy equation~\ref{eq:cipher_pool_feasibility}, and the margin cannot be replaced by $\tau$ without calibration.
\begin{table}[t]
    \centering
    \caption{Sufficient representation dimensions for the illustrative capacity-separation regimes under A1--A3.}
    \label{tab:cipher_dimension}
    \begin{tabular}{ccc}
        \toprule
        Certified margin $\delta$ & Twofold reduction ($\kappa=2$) & Tenfold reduction ($\kappa=10$)\\
        \midrule
        $0.05$ & $d\geq6{,}586$ & $d\geq7{,}872$\\
        $0.06$ & $d\geq4{,}571$ & $d\geq5{,}464$\\
        $0.07$ & $d\geq3{,}357$ & $d\geq4{,}012$\\
        \bottomrule
    \end{tabular}
\end{table}

\subsubsection{Signal retention and output fidelity}
\label{sec:cipher_fidelity}

Signal retention does not imply that clipping leaves the MoE output unchanged. To quantify a separate local fidelity property, fix a token $i$ with original routed mass $m_i>0$, normalize its native gates as $\omega_e=g_{i,e}/m_i$, and write $\vv_e=m_i f_e(\vx_i;\theta_e)$. Then $\sum_e\omega_e=1$ and the original routed output is $\sum_e\omega_e\vv_e$. Assume $\|\vv_e\|\leq V$, absorbing the original routed mass and any model-specific output scale into this bound. For a retained expert set $\mathcal U$ with mass $r=\sum_{e\in\mathcal U}\omega_e>0$, the strict output with renormalized weights satisfies
\begin{equation}
\begin{aligned}
    \left\|\sum_e\omega_e\vv_e-
    \sum_{e\in\mathcal U}\frac{\omega_e}{r}\vv_e\right\|
    &\leq V\sum_{e\notin\mathcal U}\omega_e
    +V\left(\frac1r-1\right)\sum_{e\in\mathcal U}\omega_e\\
    &=2V(1-r).
\end{aligned}
\label{eq:cipher_strict_fidelity}
\end{equation}
If $r=0$ and the routed branch returns zero, its deviation is at most $V$. For general modified weights $\omega'_e$, including reroute, define both weight vectors over the union of their supports. The triangle inequality gives
\begin{equation}
    \left\|\sum_e\omega_e\vv_e-\sum_e\omega'_e\vv_e\right\|
    \leq V\|\omega-\omega'\|_1.
    \label{eq:cipher_reroute_fidelity}
\end{equation}
Unchanged shared-expert and residual contributions cancel in this comparison. If the downstream loss is $L_{\mathrm{loss}}$-Lipschitz in the block output, its one-step change is bounded by $2L_{\mathrm{loss}}V(1-r)$ for strict and by $L_{\mathrm{loss}}V\|\omega-\omega'\|_1$ for a general weight change. These statements hold at fixed expert outputs and fixed downstream computation. They do not bound the accumulation of parameter changes during optimization. Consequently, the capacity analysis motivates affinity-aware workload reduction, while training stability, specialization, and final model capability remain separate empirical evaluations.

\subsection{Training Configurations of CIPHER-MoE}
\label{app:configuration}
\begin{table*}[h!]
\centering
\small
\caption{Training configurations for MoE models.}
\label{tab:training-config}

\resizebox{\textwidth}{!}{%
\begin{tabular}{lclrrl}
\toprule
\textbf{Model}
& \textbf{Nodes}
& \textbf{Parallel configuration}
& \textbf{Seq. length}
& \textbf{GBS}
& \textbf{MoE configuration} \\
\midrule

\textbf{DeepSeek-V4-Pro}
& 32
& TP2/PP8/EP64/CP1/ETP1/VPP4 + SP
& 4K
& 256/8192
& \shortstack[l]{384 experts; learned Top-6} \\

\textbf{DeepSeek-V4-Flash}
& 8
& TP2/PP4/EP16/CP1/ETP1/DP16 + SP
& 8K
& 128
& \shortstack[l]{256 experts; learned Top-6; sqrt-softplus gate} \\

\textbf{DeepSeek-V3}
& 16
& TP4/PP16/EP4/CP1/DP4 + SP
& 8K
& 128
& \shortstack[l]{256 experts; Top-8 routing; sigmoid gate} \\

\textbf{GLM-5}
& 16
& TP2/PP8/EP32/CP1/ETP1/VPP2/DP16/EDP1
& 8K
& 128
& \shortstack[l]{256 experts; Top-8 routing; 
  sigmoid gate} \\

\bottomrule
\end{tabular}%
}
\end{table*}

Table~\ref{tab:training-config} summarizes the training and routing configurations used in our experiments. The affinity threshold $\tau$ specifies the minimum normalized token--expert affinity required to retain a candidate assignment; assignments with affinity below this threshold are filtered out. On DeepSeek-V4-Flash, we evaluated LocMoE with $\tau\in\{0.005,0.01,0.02,0.03\}$ and found that $\tau=0.01$ provided the most favorable overall performance. We therefore adopt $\tau=0.01$ for LocMoE in all subsequent experiments to ensure a consistent configuration across models and datasets. Both variants of CIPHER-MoE use a capacity factor of $8$; CIPHER-reroute additionally uses a fill-target factor of $4$. The capacity factor denotes the capacity limit as a multiple of the uniform expert load. For locality loss, $\lambda_{loc}$ is set to $0.01$ for both LocMoE and CIPHER-MoE. These settings remain robust across different models and datasets.

\end{document}